\documentclass{article}

\usepackage{arxiv}

\usepackage[utf8]{inputenc} 
\usepackage[T1]{fontenc}    
\usepackage{hyperref}       
\usepackage{url}            
\usepackage{booktabs}       
\usepackage{amsfonts}       
\usepackage{nicefrac}       
\usepackage{microtype}      
\usepackage{lipsum}		
\usepackage{graphicx}
\usepackage{natbib}
\usepackage{doi}
\usepackage{tikz-cd}
\usepackage{mathtools}

\usepackage{txfonts} 

\usepackage{amsmath}
\usepackage{amssymb}
\usepackage[boxruled]{algorithm2e}

\newtheorem{theorem}{Theorem}
\newtheorem{definition}{Definition}

\title{On The Universality of Diagrams for Causal Inference and The Causal Reproducing Property}

\author{ Sridhar Mahadevan \thanks{Draft undergoing peer review. Comments welcome}\\
	Adobe Research and University of Massachusetts, Amherst\\
	\texttt{smahadev@adobe.com, mahadeva@umass.edu} 
}

\date{}

\begin{document} 

\maketitle

\begin{abstract} 
Humans have sought to understand causality since ancient times. 2500 years ago, Plato remarked that {\em "Everything that becomes or changes must do so owing to some cause; for nothing can come to be without a cause"}. In more modern times, Judea Pearl developed a seminal theory of causality based on directed acyclic graph (DAG) models, where a variable $X$ has a causal influence on variable $Y$ if a directed path exists between $X$ and $Y$.  We propose Universal Causality, a mathematical framework based on category theory, that generalizes both Plato and Pearl's conceptions of causality. In our framework, {\em objects are defined by the set of causal influences exerted upon them.} More formally, universal causal models are defined as categories consisting of objects and morphisms between them representing causal influences, as well as structures for carrying out interventions (experiments) and evaluating their outcomes (observations). Functors map between categories, and natural transformations map between a pair of functors across the same two categories. Abstract causal diagrams in our framework are built using universal constructions from category theory, including the {\em limit} or {\em co-limit} of an abstract causal diagram, or more generally, the Kan extension. We present two foundational results in universal causal inference. The first result, called the Universal Causality Theorem (UCT), pertains to the {\em universality of diagrams}, which are viewed as  {\em functors} mapping both objects and relationships from an indexing category of abstract causal diagrams to an actual causal model whose nodes are labeled by random variables, and edges represent functional or probabilistic relationships. UCT states that any causal inference can be represented in a canonical way as the co-limit of an abstract causal diagram of {\em representable objects}. UCT follows from a basic result in the theory of sheaves. The second result, the Causal Reproducing Property (CRP), states that any causal influence of a object $X$ on another object $Y$ is representable as a natural transformation between two abstract causal diagrams.  CRP follows from the Yoneda Lemma, one of the deepest results in category theory. The CRP property is analogous to the reproducing property in Reproducing Kernel Hilbert Spaces that served as the foundation for kernel methods in machine learning. UC enables the study of causal inference over a much richer class of models by exploiting a powerful set of construction tools from category theory, including adjunctions, braiding, decorated cospans and operads  that capture more sophisticated forms of interaction than the popular graph-based paradigm.
\end{abstract}


\section{Introduction}

\begin{figure}[h]
\centering
\begin{minipage}{0.5\textwidth}
\includegraphics[scale=0.3]{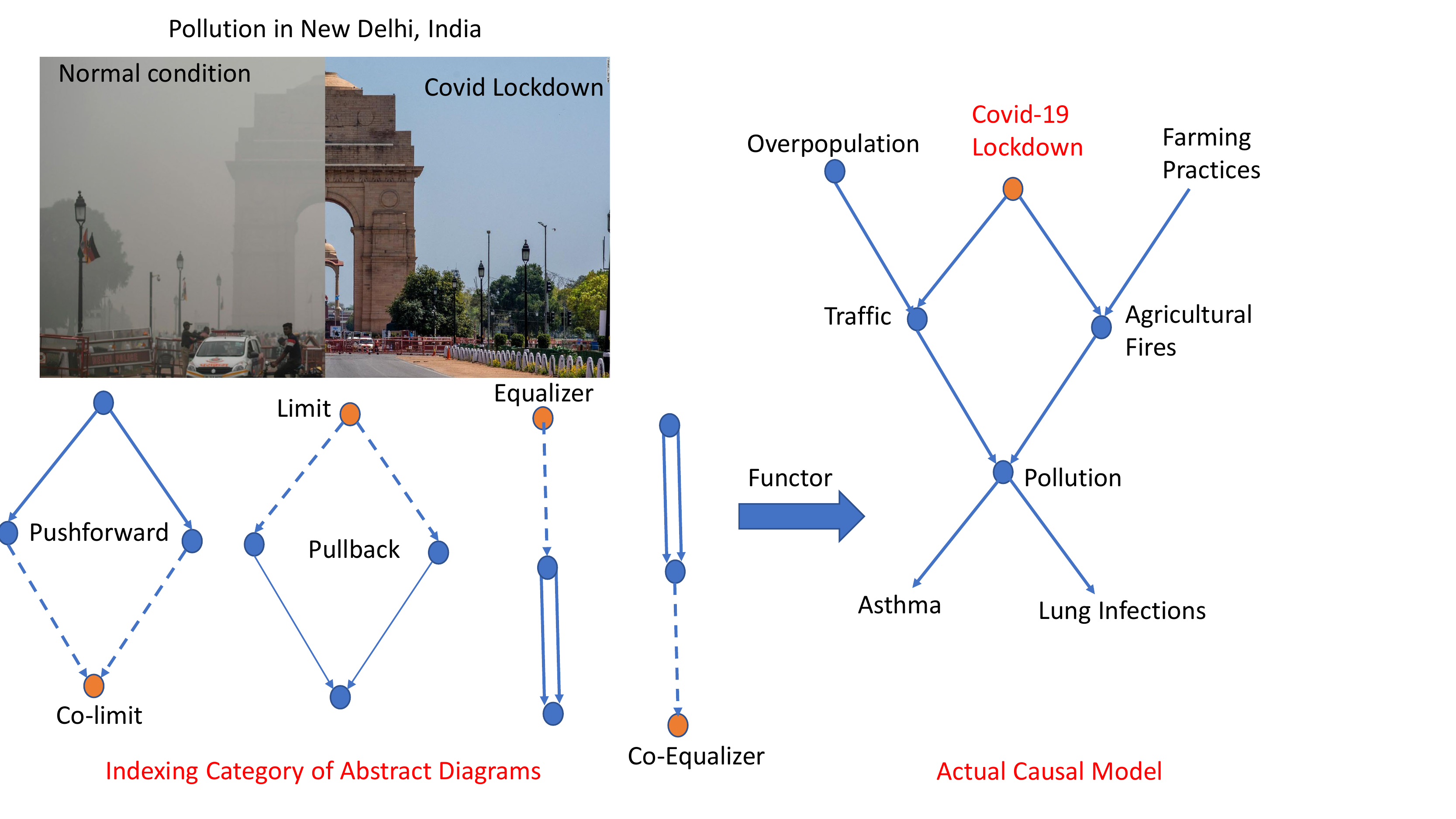}
\end{minipage}
\caption{Causality studies the question of how an object (e.g., air pollution in New Delhi, India) {\em changes} due to an intervention on another object (e.g., Covid-19 lockdown). Causal influences are transmitted along paths in a causal model. In the directed acyclic graph (DAG) model shown, causal influence corresponds to a directed path from a variable $X$ into another variable $Y$ \cite{pearl:causalitybook}.  Covid-19 lockdown caused both a reduction in traffic as well as less burning of crops, both of which dramatically reduced air pollution in New Delhi. Universal Causality (UC) formulates the problem of causal inference in an abstract category ${\cal C}$ of causal diagrams, of which DAGs are one example, where objects can represent  arbitrary entities that interact in diverse ways (see the range of possibilities in Figure~\ref{causal-cats}). Objects interact through morphisms that can represent functional \cite{sem:book}, measure-theoretic \cite{heymann:if,witsenhausen:1975}, probabilistic \cite{pearl:bnets-book} or topological relationships \cite{sm:homotopy}, with the only requirement that they be composable. The UC framework poses any causal inference question of the effect on an object $X$ by an intervention on any other object in the model as a change in the presheaf object associated with that node in the contravariant functor category $\hat{{\cal C}} = $ {\bf Set}$^{{\cal C}^{op}}$ of presheaves.  In a  directed graphical model, the set of all causal influences on a variable is simply the set of all directed paths entering the variable, and defines its presheaf functor. The Universal Causality Theorem (UCT) states any causal inference is representable as a contravariant functor object, which is a co-limit of an abstract causal diagram functorially mapping a small indexing category of diagrams to an actual causal model. This result follows from the theory of sheaves (see Proposition 1, page 41, in \cite{maclane:sheaves}), and establishes the primacy of diagrams in causal inference.  The Causal Reproducing Property (CRP) states that presheaf objects act as the principal ``representers" of all causal information, namely {\bf Hom}$_{\cal C}(X,Y) \simeq$ {\bf Nat}({\bf Hom}$_{\cal C}(-, X)$,{\bf Hom}$_{\cal C}(-, Y))$. CRP follows directly from the Yoneda Lemma \cite{maclane:71}, and is analogous to the reproducing property in Reproducing Kernel Hilbert Spaces that defined kernel methods in machine learning \cite{kernelbook}.}   
\label{uc2}
\end{figure}

Humans have sought causal explanation since ancient times.  The Greek philosopher \citet{plato} said that {\em "Everything that becomes or changes must do so owing to some cause; for nothing can come to be without a cause"}. For a more modern perspective, Pearl's seminal work on causal inference modeled  causal influences using directed acyclic graph (DAG): ``a variable $X$ is said to have a causal influence on a variable $Y$ if a directed path from $X$ to $Y$ exists in every minimal structure consistent with the data" (Definition 2.3.1, page 45 in \cite{pearl:causalitybook}). We propose a  framework called Universal Causality (UC) (see Figure~\ref{uc2}), which can be viewed as a mathematical abstraction of both Plato and Pearl's conceptualization of causal inference: {\bf every object in UC is represented as an amalgam of all possible causal influences upon it}. More precisely,  UC draws its power from {\em category theory} \cite{maclane:71,riehl2017category,goldblatt:topos,Johnstone:592033} and the theory of sheaves \cite{maclane:sheaves}, both of which are singularly well-suited to study the problem of causal inference. Category theory is the science of building and reasoning about complex systems \cite{fong2018seven}. Sheaf theory \cite{maclane:sheaves} can be viewed as the abstract mathematical study of ``data structures": how to build a globally consistent model by ``collating" local pieces. The rich body of work in computer science, AI and statistics on causal inference \cite{pearl:bnets-book,pearl:causalitybook,spirtes:book} has developed seminal insights into causal inference, which have a natural, but so far untapped, synergy with the major themes of category theory and sheaf theory. 

The {\bf Yoneda Lemma} states that any object in a category is defined up to isomorphism in terms of the interactions it makes with all other objects. An analogy from particle physics proposed by Theo Johnson-Freyd might help give insight into the Yoneda Lemma: ``You work at a particle accelerator. You want to understand some particle. All you can do is throw other particles at it and see what happens. If you understand how your mystery particle responds to all possible test particles at all possible test energies, then you know everything there is to know about your mystery particle". The Yoneda Lemma states that the set of all morphisms into an object $X$ in a category ${\cal C}$, denoted as {\bf Hom}$_{\cal C}(-,X)$ and called the {\em contravariant functor} (or presheaf),  is sufficient to define $X$ up to isomorphism. Accordingly, any change in $X$ due to a causal experiment must be accompanied by a change in the presheaf contravariant functor object.  Thus, the Yoneda Lemma can be seen as a mathematical way to formalize both Plato's and Pearl's insight into causality. The universal property of causality, which defines the proposed UC framework, formulates causal inference in the functor category of presheaves $\hat{{\cal C}} = $ {\bf Set}$^{{\cal C}^{op}}$.

\begin{figure}[h]
\centering
\begin{minipage}{0.5\textwidth}
\includegraphics[scale=0.3]{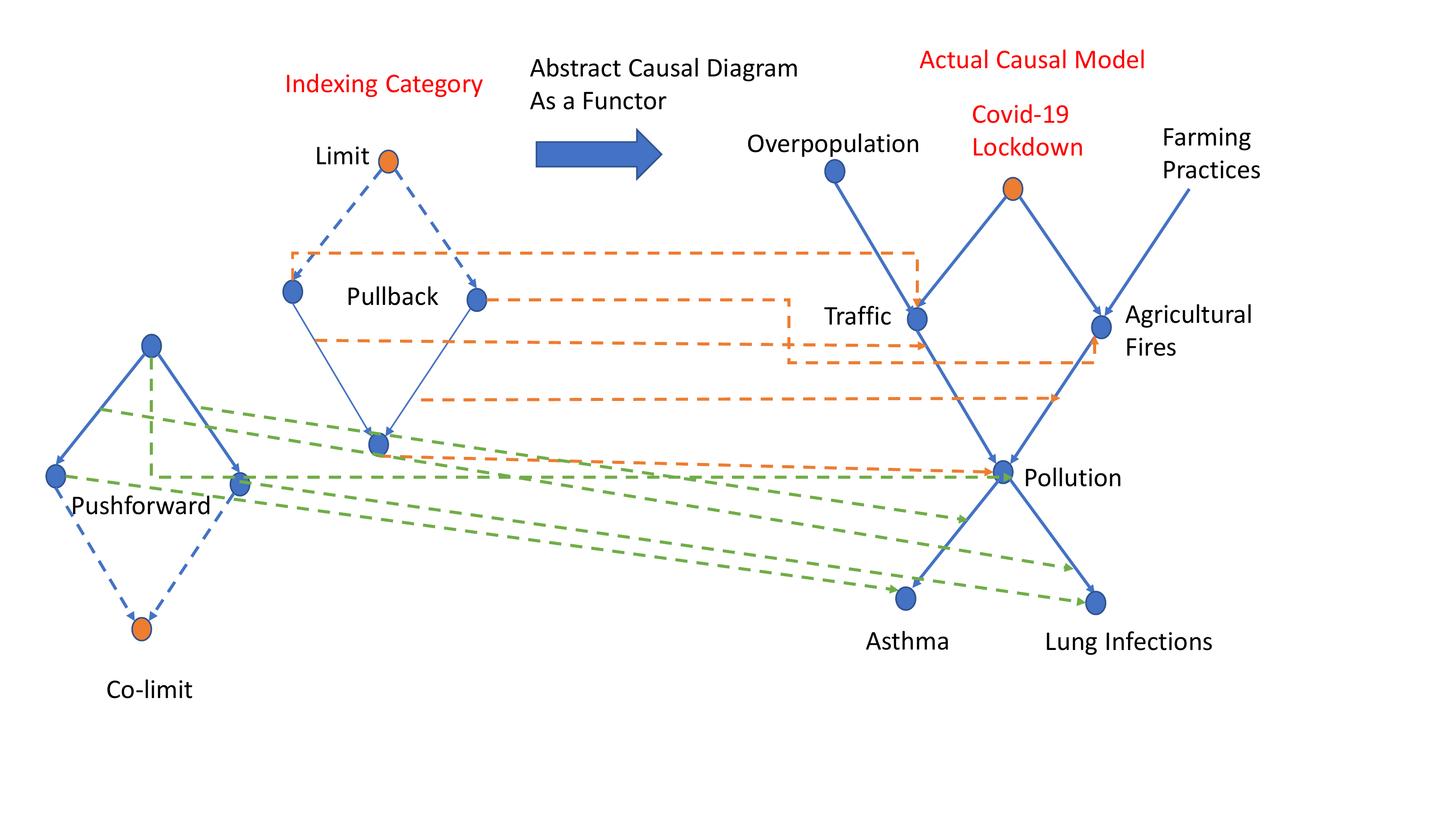}
\end{minipage} \vskip -0.3in
\caption{The notion of diagram in Universal Causality is more abstract than previous diagrammatic representations in causal inference. For example, in this causal DAG model, {\bf Pollution} is a collider node between the two causes {\bf Traffic} and {\bf Agricultural Fires}. In our framework, an abstract causal diagram is a functorial mapping between an indexing category of diagrams to the actual causal model. Thus, the abstract diagram on the left maps functorially into the actual causal model by mapping each object $\bullet$ into a causal variable in the model, and each morphism into an edge in the DAG. .  Universal Causal Models are defined using universal constructions, such as pullbacks, pushouts, co-kernels and kernels, which are all special cases of {\em limit} or {\em co-limits}. For example, the limit of the abstract causal diagram in this example is a common cause $Z$ of {\bf Traffic} and {\bf Agricultural Fires}, such that $Z$ satisfies a universal property, namely every other common cause $W$ must factor through it. Although {\bf Covid-19} is certainly a common cause of {\bf Traffic} and {\bf Agricultural Fires} both being significantly lower than normal, in general, it might not be the limit, as it is possible there might be many other common causes (e.g., a weather event). Dually, the co-limit of {\bf Asthma} and {\bf Lung infection} is some common effect $E$ such that every other effect of these conditions must factor through $E$. Co-limits and limits generalize notions such as disjoint unions and products in sets, and joins and meets in partial orders.}    
\label{uc-diagram}
\end{figure}

The fundamental theoretical insight provided by sheaf theory is on the {\em universality of diagrams} in causal inference. More formally, any causal inference in UC is defined as an object in the contravariant functor category {\bf Set}$^{{\cal C}^{op}}$ of presheaves, which is representable as the co-limit of a small indexing diagram that serves as its universal element. The notion of diagram in UC is more abstract than previous diagrammatic representations in causal inference (see Figure~\ref{uc-diagram}). For example, in causal inference using DAGs, a collider node $B$ defines the structure $A \rightarrow B \leftarrow C$. In our framework, an abstract causal diagram functorially maps from an indexing category of diagrams, such as $\bullet \rightarrow \bullet \leftarrow \bullet$ into the actual causal diagram $A \rightarrow B \leftarrow C$. Abstract causal diagrams themselves form a category of functors.  The fundamental notion of a {\em limit} or {\em co-limit} of a causal diagram in UC, are abstract versions of more specialized {\em universal constructions}, such as ``pullbacks", ``pushouts", ``co-equalizers" and ``equalizers", commonly used in category theory, but hitherto not been studied in causal inference. For example, the limit of the abstract causal diagram mapping $\bullet \rightarrow \bullet \leftarrow \bullet$ into the actual causal model $A \rightarrow B \leftarrow C$ is defined as a variable, say $Z$, that acts as a ``common" cause between $A$ and $C$ in a ``universal" manner, in that any other common cause of $A$ and $C$ must factor through $Z$. In fact, a rich repertoire of other construction tools have been developed in category theory, including Galois extensions, adjunctions, decorated cospans, operads, and props, all of which enable building richer causal representations than have been previously explored in the causal inference literature. 

UC is representation agnostic: it applies to graph-based representations equally well as it does to less familiar representations for causal inference, including measure-theoretic information fields  representations that generalize Pearl's do-calculus beyond graphs \cite{heymann:if,witsenhausen:1975} as well as finite space topological representations \cite{heymann:if,sm:homotopy}.  Logic has long been a popular language for codifying causal inference in philosophy. \citet{DBLP:journals/jphil/Lewis73} proposed a well-known theory of counterfactuals, which analyzes statements such as ``If Kangaroos have no tails, they would fall over". To interpret this counterfactual statement, Lewis proposes that counterfactual propositions are evaluated in terms of their truth value in all worlds that are similar to ours that fall into an equivalence class (e.g., where kangaroos have no tails, but everything else remains the same). \citet{halpern:ac} formulates the problem of actual causality in terms of logic.  A topos is a category that captures the essential aspects of the category of sets, but it is more general than sets, and admits a form of intuitionistic logic. Counterfactual reasoning of the type that was made popular by Lewis \cite{DBLP:journals/jphil/Lewis73} can be modeled as reasoning in a topos \cite{topos-counterfactual}. 

\begin{figure}[h] 
\centering
\caption{Universal Causality (UC) uses category theory to characterize the universal property of causal inference that applies across a diverse landscape of representations shown in the table below. \label{causal-cats}}
\vskip 0.1in
\begin{minipage}{0.7\textwidth}
\vskip 0.1in
\includegraphics[scale=0.4]{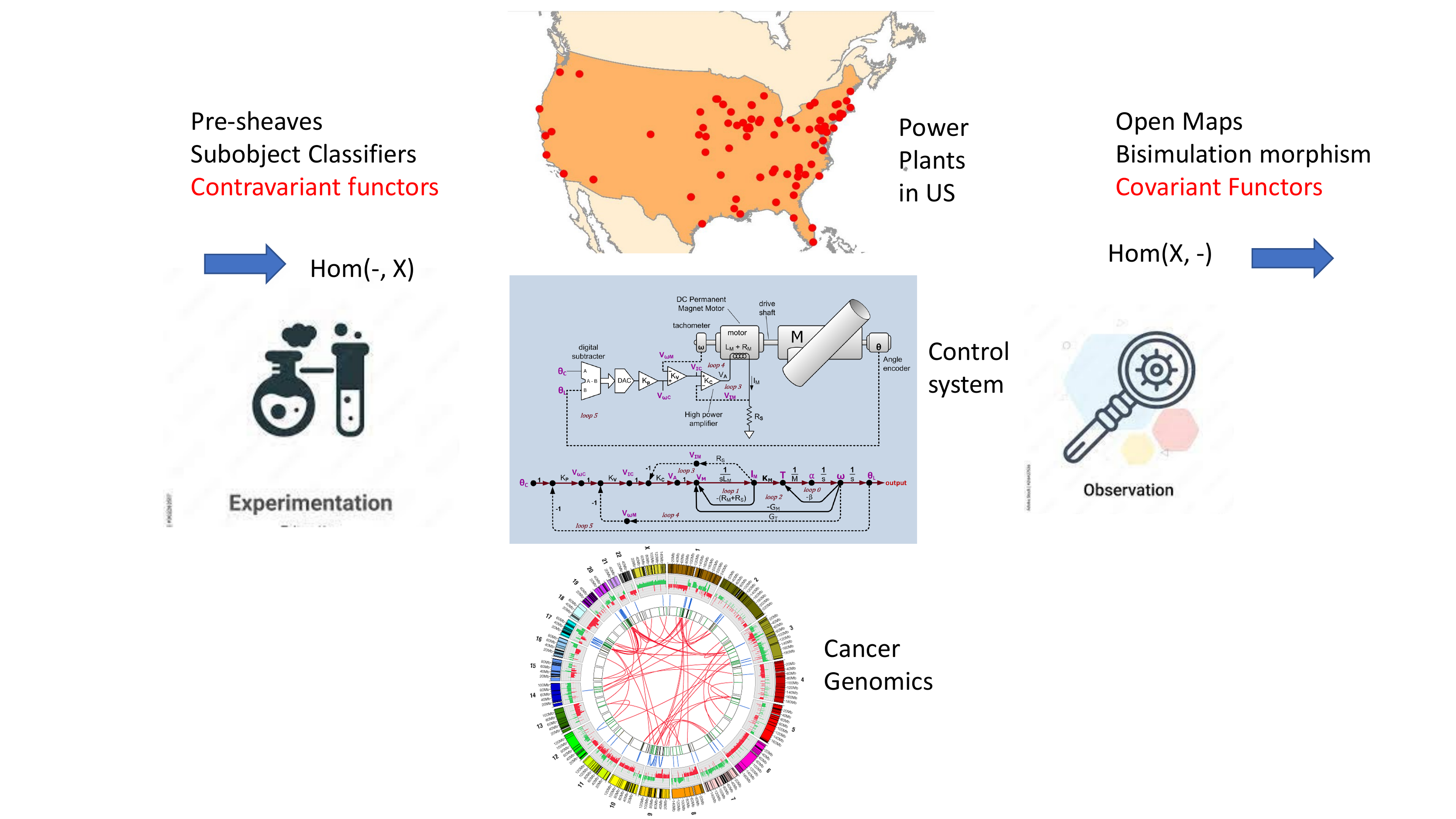}
 \begin{small}\hfill 
  \begin{tabular}{|c|c|c|c|} \hline 
Representation & Objects  & Morphisms & References  \\ \hline 
Graphs & Variables $X,Y,\ldots$ & Paths & \cite{pearl-book} \\ \hline 
Topological Spaces & Open sets $\{\{a\}, \{ b \}, \{a,b\}\}$  & Fences & \cite{barmak} \\ \hline  
Information fields & Measurable Spaces & Measurable mappings & \cite{witsenhausen:1975} \\ \hline
Resource Models & Monoidal resources & Profunctors & \cite{fong2018seven} \\ \hline
Concurrent Systems & Program variables & Bisimulation morphisms & \cite{JOYAL1996164} \\ \hline
Dynamical systems & States & Processes & \cite{sm:udm} \\ \hline
Counterfactuals & Propositions & Proofs & \cite{DBLP:journals/jphil/Lewis73} \\ \hline 
Network Economy & Consumers/Producers & Trade & \cite{nagurney:vibook} \\ \hline 
Discourse Sheaves & Users & Communication & \cite{discourse-sheaves} \\ \hline
\end{tabular}
\end{small}
\end{minipage}
\end{figure}

Even a cursory study of the past literature on causal inference would immediately overwhelm the reader, since a veritable ``Tower of Babel" repertoire of languages and representations have been proposed to understand causality. UC can be viewed as a ``Rosetta Stone" \cite{Baez_2010} to translate across diverse representations used in causal inference (see  Figure~\ref{causal-cats}).  Our paper is by no means the first paper to use category theory to study causal inference. For example, \citet{fong:ms} explored the use of symmetric monoidal categories as a language for defining Bayesian networks, a popular probabilistic graphical model for defining causal models. However, our approach differs in its application of ideas from sheaf theory and Yoneda Lemma, as well as using universal construction tools, such as the Kan extension, to formulate central results, such as establishing the universality of abstract causal diagrams, and the causal representer property.

Reproducing Kernel Hilbert Spaces (RKHS's)  transformed the study of machine learning, precisely because they are the unique subcategory in the category of all Hilbert spaces that have representers of evaluation defined by a kernel matrix $K(x,y)$ \cite{kernelbook}. The reproducing property in an RKHS is defined as $\langle K(x, -), K(-, y) \rangle = K(x,y)$. An analogous but far more general reproducing property holds in the UC framework, based on the Yoneda Lemma. The {\bf causal reproducing property} (CRP) states that the set of all causal influences between two objects $X$ and $Y$ in a category ${\cal C}$ can be built out of its presheaf functor objects, that is {\bf Hom}$_{\cal C}(X,Y) \simeq$ {\bf Nat}({\bf Hom}$_{\cal C}(-, X)$,{\bf Hom}$_{\cal C}(-, Y))$.  In other words, presheaves act as ``representers" of causal information, precisely analogous to how kernel matrices act as representers in an RKHS.

Let us make this abstract discussion more concrete in the setting of causal directed acyclic graph (DAG) models, which have been extensively studied in causal inference for the past several decades \cite{pearl-book,spirtes:book,hedge}. The fundamental concept of $d$-separation states that a random variable $X$ (represented as a vertex in a DAG $G$) is  $d$-separated from another variable $Y$ conditional on an observation set $Z$ if and only if every (undirected) path from the vertex $X$ to $Y$ satisfies a certain property: if the path has no collider node, meaning the path lacks any triple of nodes with the pattern $u \rightarrow v \leftarrow z$, then some node $w$ in every such path must be in the conditioning set $w \in Z$. Alternatively, if the path has a collider $u \rightarrow v \leftarrow z$, then neither $v$ nor any of its descendants must be in $Z$. What is fundamentally important here is that causal inference in the category of DAG models involves a functorial mapping into the category of {\bf Set}, in fact precisely objects in the presheaf of all paths into {\cal Hom}$_{\cal G}(-, Y)$ (which satisfy the $d$-separation property with respect to $X$ and $Z$). What is even more crucial is the reverse implication that follows from the fundamental theorem of Universal Causality: any causal inference that is done by constructing such presheaf contravariant functor objects must be representable as the (co)limit of a small diagram in the category of diagrams. In other words, it is this reverse implication that is of crucial importance in understanding UC. Of course, it is relatively straightforward to realize that all paths in the DAG from $X$ to $Y$ form a presheaf covariant functor. What is not obvious is that any such presheaf object must necessarily be representable by a causal diagram. That is the core novelty of the UC framework in being able to show why diagrams in causal inference reflect a universal property. We aim to show that UC not only provides novel theoretical insights into causal inference, but also yields algorithmic insights for scaling to real-world problems.

Causal inference in the real world is often made significantly more challenging due to unobservable confounders \cite{pearl-book}. In UC, the implication of this problem is that the presheaf functor object {\cal Hom}$_{\cal C}(-, X)$ cannot be computed as morphisms from unobservable confounders are not available.  Every concept in category theory can be seen as a Kan extension, including the Yoneda Lemma as well as co-limits and limits, and thus, the Kan extension emerges as a unifying motif in UC to pose a variety of problems in causal inference. The intuitive idea behind the Kan extension is to approximate a desired functor $F: {\cal C} \rightarrow {\cal E}$ along the direction of an available functor $G: {\cal C} \rightarrow {\cal D}$. In effect, this construction {\em extends} the scope of the original functor $F$ from the category ${\cal C}$ to the category ${\cal D}$. In applying Kan extensions to causal inference over unobserved confounders, ${\cal C}$ is the full category of all causal objects, and ${\cal D}$ is the subcategory of observable causal objects. 

\section{Background and Related Work} 

Universal Causality differs from the vast literature of previous work in the field based on one very simple, but important, idea: in UC, causality is viewed as a {\em functor} that maps from some indexing category of abstract causal diagrams into an actual causal model. We review some background material in this (long) section, emphasizing the places where UC differs from previous approaches. In a nutshell, UC emphasizes the synergy between causality and category theory, which in our view has been underappreciated.  Category theory provides an exceptionally rich language to build complex systems, but has not directly embraced ideas from causality: how do objects {\em change} in a category when a causal intervention experiment is done that changes some other object? For example, \citet{fong2018seven} have written a wonderful book on applied category theory, detailing a rich set of construction methods using symmetric monoidal categories to build complex models, but they do not mention the topic of causal intervention in the book. On the other hand, causal inference has long used representations, such as directed graphs, probability theory, and statistics, but has not exploited the deeper ideas of category theory, such as the Yoneda Lemma, the topological structure of sheaves, and the rich set of construction tools in symmetric monoidal categories. Our paper aims to bring these interlocking ideas closer together, which necessitates more than the usual amount of tutorial material to make the paper accessible to both communities. Readers familiar with causal inference or category theory can skim the appropriate sections, although we do introduce novel perspectives of these familiar topics even in this background review (for example, limits and colimits of abstract causal diagrams, and viewing graphical models in terms of {\em sheafs} \cite{maclane:sheaves}). 

\subsection{Causality} 

Causality seems to have fascinated mankind from the earliest periods of our history. Besides the Platonic concept of causality as defining an object, from around the same time period, \citet{aristotle}  defined the notion of an {\em efficient cause} as "primary source of the change", or an object whose activity results in an effect on another object. In more modern philosophy, \citet{descartes} divided causes into {\em particular causes} (e.g., laws of motion) and {\em general causes} (e.g., God). \citet{hume} pioneered the notion of inferring causality from experience, stating that {" "There are no objects, which by the mere survey, without consulting experience, we can determine to be the causes of any other; and no objects, which we can certainly determine in the same manner not to be the causes"}.  In 1876, writing in the preface of Charles Darwin's book on treatment of plant varieties \cite{darwin:1876}, the statistician Frances Galton pioneered the use of statistics to quantify a causal experiment: 
\begin{quote}
The observations. . . have no prima facie appearance of regularity. But as soon
as we arrange them in order of their magnitudes,. . . . We now see, with few exceptions,
that. . . the largest plant on the crossed side. . . exceeds the largest plant on the self-fertilised
side, that. . . the second exceeds the second,. . . and so on. . . ”.
\end{quote}
The modern 20$^{th}$ century study of causality was shaped by statisticians, such as Fisher who pioneered the concept of a randomized clinical trial (RCT) (now called A/B testing and widely used by businesses to find improvements in products and services) \cite{fisher}. By randomizing an experiment, Fisher showed that spurious correlations between potential causes and effects could be reduced. Diagrammatic representations have long been popular in causal inference, beginning more than 100 years ago with the pioneering work of geneticist Sewall Wright, who introduced the method of path coefficients \cite{Wright1921CorrelationAndCausation} to study causation in biology. More recently, causal diagrams have been seen a resurgence of popularity, due in large part to Judea Pearl's seminal work on structural causal models and the do-calcululus of causal interventions \citep{pearl:causalitybook}, which has developed into a large and fertile ground for studying causal inference in many fields \cite{spirtes:book,mdag,hedge}.
We describe Pearl's work below in more depth as it will play an important role in the UC framework as well, in particular as we show later, UC enables generalizing many aspects of this approach to category theory. 

\begin{figure}[h]
\centering
\begin{minipage}{0.3\textwidth}
 \begin{tikzcd}[column sep=small]
& X \arrow[dl] \arrow[dr] & \\
  Z \arrow{rr} &                         & Y
\end{tikzcd}

 \begin{tikzcd}[column sep=small]
& \bullet \arrow[dl] \arrow[dr] & \\
  \bullet \arrow{rr} &                         & \bullet 
\end{tikzcd}
\end{minipage} 
\caption{In Universal Causality, causal diagrams, such as the DAG shown on top,  are themselves viewed as the co-domain of an {\em abstract causal diagram} from an indexing category ${\cal J}$, where ${\cal J}$ contains diagrams such as the one shown below. Note that the notion of a diagram is functorial, meaning that the diagram $F: {\cal J} \rightarrow {\cal C}$, where ${\cal C}$ is the actual causal model shown above labeled with $X$, $Y$, and $Z$, and the abstract diagram shown below is an element of an indexing category ${\cal J}$. \label{acd}}
\end{figure}
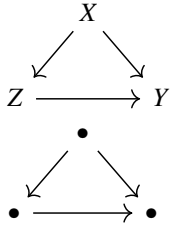

Our interpretation of the large amount of research following Pearl's framework \cite{pearl:bnets-book,pearl:causalitybook,spirtes:book} is that it fundamentally involve ideas from category theory and sheaf theory, but so far, they have remained implicit, and our goal is to make them explicit (see Figure~\ref{acd}).   For example, much work in causal graphical models explores the causal effect of intervening on an variable $Z$ in the model shown in Figure~\ref{acd}, and evaluating its impact on $Y$. Since $Y$ depends on $X$ and $Z$, Pearl's ``backdoor criterion" \cite{pearl:causalitybook} stipulates that to obtain a true estimate of the causal influence of $Z$ on $Y$, we must eliminate the ``back door" path that leads out of $Z$ to into $Y$ through $X$. One way is to intervene on $X$ (for example, by setting it to $X = 1$). The core of the UC framework is based on studying the functorial mapping from abstract causal diagrams to actual causal diagrams.

Consider a causal model represented as a directed acyclic graph (DAG) structure ${\cal G}_n = (V_n, E_n)$, for example a Bayesian network \cite{pearl:bnets-book}, where $V_n = [n] = \{1, \ldots, n\}$. The edge set $E_n$ is a set $\{(x,y) \in V_n \times V_n \ |  \ (x, y) \in E_n \}$, where $x, y \in [n]$, of comparable pairs. DAGs are constrained so that $(x, y) \in E_n$ implies $(y,x) \notin E_n$, and there are no directed cycles. A directed path $x_i, x_{i+1}, \ldots, x_j$ between variables $x_i, x_j \in [n]$ is such that each pair of nodes $(x_k, x_{k+1}) \in E$ on this path. We can define a partial ordering ${\cal P}_n$ on the vertices $[n]$ of $G_n$ from the transitive closure of the original graph ${\cal G}_n$. 

Fundamental to Pearl's approach to causal inference is defining notions like conditional independence and d-separation on the set of all (undirected) paths through a DAG. This very step, of taking a DAG and looking at the set of all paths for those satisfying some property, is an example of a functorial mapping from one category (e.g., all DAGs) to another category (the category of {\bf Set}). Bayesian networks \cite{pearl:bnets-book} and structural causal models (SCMs) \cite{pearl:causalitybook} are both used extensively in the literature on causal inference in AI and machine learning. Bayesian networks are defined by probability distributions $P_{{\cal G}_n}$ on a DAG $G_n$, such that the conditional independences defined on the graph are exactly those satisfied by the probability distribution $P_{{\cal G}_n}$ \cite{lauritzen:text}. Given a DAG $G_n$, a set of nodes $X$ is conditionally independent of a set $Y$, given a third set $Z$, where $X, Y, Z \subset [n]$, denoted $(X \Perp Y | Z)_{{\cal G}_n}$, if each directed path from a node $x \in X$ to a node $y \in Y$ is such that for any collider node $i \rightarrow j \leftarrow k$ in the path, neither $j$ nor any of its descendants $\in Z$, and there exists some non-collider node $i \rightarrow j \rightarrow k$ in the path such that $j \in Z$. In the distribution $P$, the equivalent conditional independence property is defined as $(X \Perp Y | Z)_P$ if $P(X | Y, Z) = P(X | Z)$. 

To make the connection between the Pearl's causal inference on graphs paradigm and category theory more precise, we can view a collider $i \rightarrow j \leftarrow k$ as an instance of an abstract diagram, a functor that maps from an indexing category of diagrams, such as $\bullet \rightarrow \bullet \leftarrow \bullet$ to a collider. One of the interesting benefits of doing this abstraction of colliders is that one can now ask interesting questions about universal properties: for example, given any collider node, what is its limit? This question has not been studied before in the literature on graphical models, because it requires defining the very notion of universal property, a topic not central to causal inference or graphical models in AI, but in fact, it is central to category theory.

A detailed analysis of conditional independence properties are given in \cite{lauritzen:text,pearl:bnets-book}. In SCMs, the modeling framework is generalized from a classic Bayesian network in that an SCM ${\cal M} = (U,V,f, P)$ is specified by a set of endogenous variables $V$ internal to the model, a set of exogenous variables $U$ external to the model (meaning these have no parents from within the model) ,  a set of arbitrary functions $f$ that compute the value of each endogenous variable from its ``parents" (which can include exogenous variables and other endogenous variables), and a probability distribution $P$ on exogenous variables. It is very straightforward to encode any SCM as a category, but the entire theoretical machinery of category theory, as we show later in the paper, makes it possible to define models that are far richer than SCMs that have been proposed previously in the causal inference literature. For example, causal information fields \cite{heymann:if} generalize Pearl's theory of d-separation to more complex models, where objects are measurable spaces, and edges in the model correspond to measurable functions. \citet{heymann:if} show that the rules of do-calculus can be succinctly summarized in one equation. It is also fairly easy to show that any causal information field defines a category \cite{sm:udm}, and thereby the framework of UC directly applies to causal information fields as well. 

Causal discovery from data is made more challenging by the realization that observations can only determine an equivalence class of graphical models, and fully determining directional relationships between variables requires interventional experiments. Many variants of DAGs have been developed in recognition of this issue, from chain graphs \cite{lauritzen:chain} to marginalized DAGs \cite{mdag} and hyper-edge directed graphical models \cite{hedge}. Causal discovery of a DAG model, or any of the other variants, requires {\em causal interventions} \cite{pearl:causalitybook}, where a node $i \in [n]$ in the DAG that is intervened on has all its incoming edges  deleted. A large number of studies have been conducted on constructing intervention sets \cite{DBLP:conf/uai/Eberhardt08,DBLP:journals/jmlr/HauserB12,DBLP:conf/nips/KocaogluSB17,MAOCHENG198415,prasad:aaai21}.

  In a structural causal model \cite{pearl:bnets-book}, the variables divided into two disjoint subsets $\{X_1, \ldots, X_n \} = U \sqcup V$, where $U$ are the ``exogenous" variables that have no (observable) parents in the model, and $V$ are the endogenous variables whose values are computed by ``local" functions $f_i$ of other endogenous and exogenous variables. It is straightforward to define a category of structural causal models ${\cal C}_{SCM}$, where each (endogenous or exogenous) variable is an object, and a morphism $f_i: X \rightarrow Y$ exists if the value of $Y$ depends on the value of $X$. As we will see later, encoding SCMs as a category enables using powerful theoretical tools from category theory (see review below) to analyze their structure. The probability distribution in a structural causal model is defined over the exogenous variables $P(U)$, such that for any particular configuration $u \in U$ of the exogenous variables, the values of the endogenous variables in $V$ can be uniquely determined.

\begin{definition}
Given a DAG ${\cal G}_n = ([n], E_n)$, for pairwise disjoint $X, Y, Z \subseteq [n]$, $Z$ is an {\bf adjustment set} \cite{DBLP:conf/uai/ShpitserVR10} relative to $(X, Y)$ if and only if 
\begin{equation} 
P(Y | do(X))  = \sum_{Z' \in Z} P(Y, X, Z') P(Z') 
\end{equation} 
\end{definition}
For example, the set of parents $\mbox{Pa}(X_i)$ forms an adjustment set for $(X_i,Y)$. Another adjustment set can be found from the {\em backdoor criterion} \cite{pearl:causalitybook}, which is a set of variables $Z$ such that no variable in $Z$ is a descendant of any variable in $X$, and $Z$ d-separates all paths from $X$ and $Y$ that contain an arrow into $X$. Notice how all these definitions for causal DAGs involve analyzing sets of paths into a variable $X$, which in category-theoretic terms, are subobjects of the presheaf functor {\bf Hom}$_{\cal C}(-, X)$. 
\begin{definition}
Given a DAG ${\cal G}_n = ([n], E_n)$, an ordered pair of subsets $X, Y \subseteq [n], X \cap Y = \emptyset$ is {\bf not confounded} if and only if $P(Y = y | do(X = x)) = P(Y = y | X = x)$ over all $x,y$. 
\end{definition}
In practice, most applications of causal inference in clinical trials, for example, involve confounding. Evaluating a clinical trial of a Covid-19 vaccine involves selecting a set of subjects, some of whom are given the vaccine, but the other control group are not. The problem inherently involves missing data \cite{rubin-book}: we cannot observe how patients who are not given the drug (the control group) would have fared had they been given the drug, and vice versa, we cannot determine for patients who were given the vaccine, whether they would have contracted the virus if they had not been given the vaccine. In the UC framework, confounding implies that the presheaf functor {\bf Hom}$_{\cal C}(-, X)$ cannot be fully determined as paths through unobservable variables cannot be evaluated the same way as paths through observed variables. Techniques like Kan extensions \cite{maclane:71} can be used to approximate functors  $F: {\cal C} \rightarrow {\cal E}$ along some other functor $K: {\cal C} \rightarrow {\cal K}$, and provide a novel way to formulate confounded causal inference.

In the Fisher randomized clinical trials tradition, perhaps the best known work on causal inference is that of Rubin's ``potential outcomes" \cite{rubin-book}. 

\begin{definition}
In the Neyman-Rubin {\bf potential outcomes} framework \cite{rubin-book}, causal effects are modeled using a binary treatment indicator $T_i$ for each unit $i = 1,\ldots, n$, where $T_i = 1$ signifies that unit $i$ receives a (binary) treatment, and $T_i = 0$ denotes unit $i$ is a control.  The {\bf potential outcome} $Y_i = T_i Y_i(1) + (1 - T_i)Y_i(0)$. Confounding is modeled in the Neyman-Rubin framework by strong ignorability: $\{ Y_i(0), Y_i(1) \} \Perp X | Z $, for some set of covariates $X$. Strong ignorablity implies that $Z$ satisfies the back-door criterion \cite{pearl:causalitybook}. 
\end{definition}
A classic measure of the effect of treatment on a unit's potential outcome is {\em average treatment effect} (ATE), which is simply defined as
\begin{equation}
\label{ATE}
ATE = \mathbb{E}(Y(1)) - \mathbb{E}(Y(0)) = \mathbb{E}(Y | do(X) = 1) - \mathbb{E}(Y | do(X) = 0) 
\end{equation}. 
The stable unit treatment value assumption (SUTVA) \cite{rubin-book} is typically assumed, which implies that $Y(1) \Perp Y(0) | Z$, where $Z$ is an adjustment set satisfying the backdoor condition. One of the most widely used algorithms for estimating ATE is the classic Horvitz-Thompson (HT) method, developed in the mid 1950s \cite{DBLP:reference/stat/Maiti11}. The HT estimator has been the subject of numerous studies over many decades characterizing its bias and variance. Recently, an interesting Gram-Schmidt random walk algorithm \cite{gs-random-walk} has been proposed based on the HT estimator, which builds on work in theoretical computer science on discrepancy theory \cite{DBLP:conf/stoc/BansalJ0S20}.

\subsection{Category Theory} 

\begin{figure}[h]
\centering
\begin{minipage}{0.75\textwidth}
\includegraphics[scale=0.5]{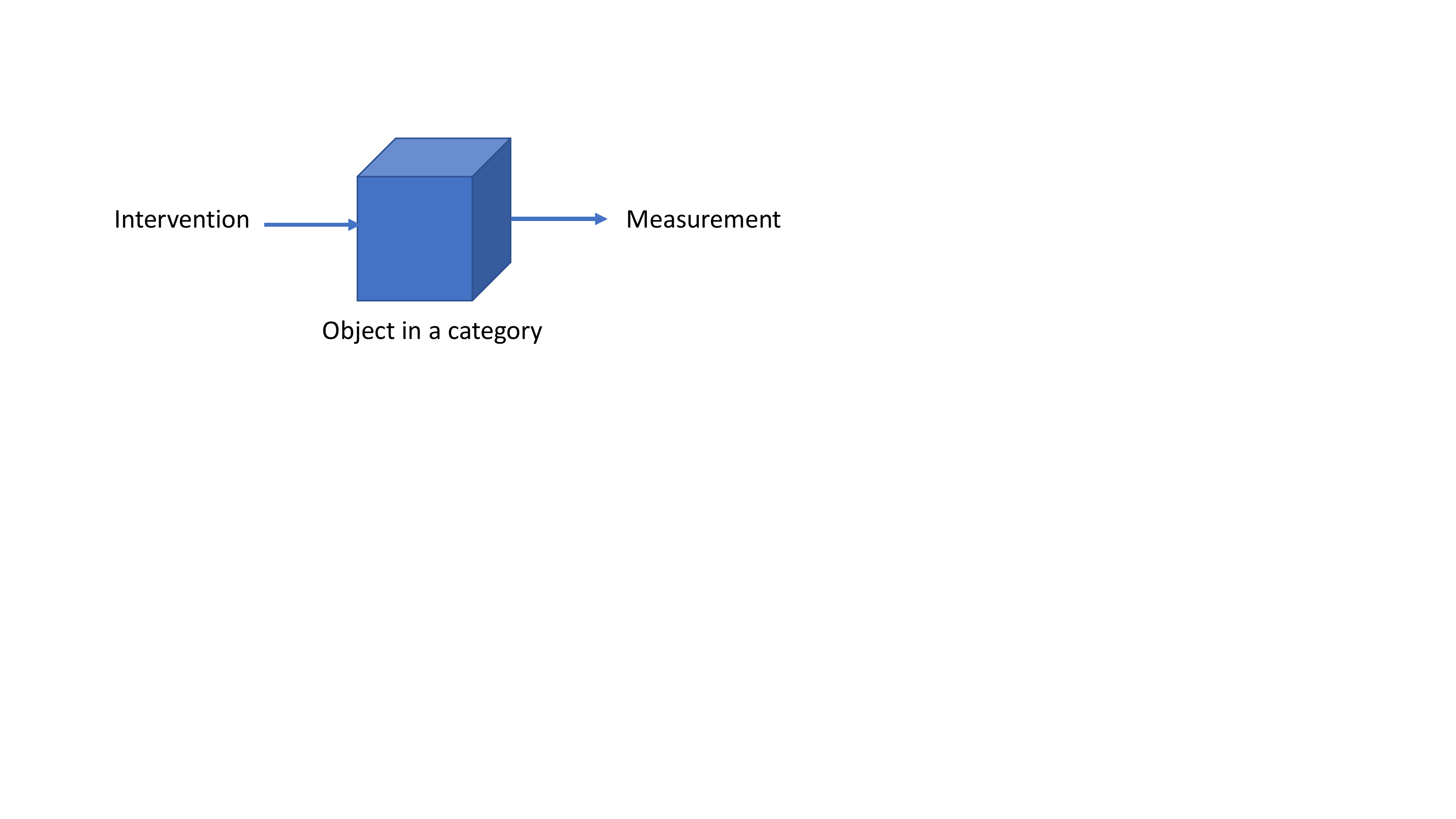}
\end{minipage} \vskip -2in
\caption{Category theory differs fundamentally from set theory in treating objects as ``black boxes": this intrinsic property makes category theory singularly useful as a way of formulating causal inference. An intervention in a causal experiment can be formulated as {\em selecting} a value of a variable, which in category theory corresponds to a morphism {\em into} an object. A {\em measurement} in contrast refers to a morphism {\em out} of an object. The set of all interventions on the objects in a category ${\cal C}$ can be represented as the {\em comma} category ${\cal I} = 1 \downarrow {\cal C}$, whose objects are morphisms from the unit object {\bf 1} to any object $c \in {\cal C}$, and whose morphisms are $i_f: ({\bf 1} \rightarrow c) \rightarrow ({\bf 1} \rightarrow d)$, where $f: c \rightarrow d$ is a morphism in $c$. \label{comma-cat}} 
\label{uc}
\end{figure}

Category theory is built from two simple components, {\em objects} and {\em morphisms}. The breadth of its utility (as witnessed by the table in Figure~\ref{causal-cats}) reflects the widespread use of these notions across science and engineering. More important to its application to causal inference is that category has a {\em built-in} notion of intervention (see Figure~\ref{uc}). A set is described by listing its contents. An object in a category, even if it is a set, is a ``black box", and accessible only through interventions that select an ``element" from the object, and that ``transmit" its value to another object as a ``measurement".  Category theory is based on universal constructions. We will explain the concept of universality in category theory with a simple example. Consider the Cartesian product of two sets $X \times Y =\{(x, y) | x \in X, y \in Y\}$, which is defined in set theory as the set of all ordered pairs (e.g., the Euclidean plane $\mathbb{R}^2$ is defined from the real numbers $\mathbb{R}$ using the Cartesian product). Although this ``definition" specifies ``what" the Cartesian product does, it does not explain ``why" it is unique.  In category theory, one instead asks: what is the underlying {\em universal property} that defines the Cartesian product? The answer comes not from listing ordered pairs, but by producing a {\em universal element} $X \times Y$ denoting the ``Cartesian product" object, and a pair of mappings $p$ and $q$ such that $X \xleftarrow[]{p} X \times Y \xrightarrow[]{q} Y$, where $p$ and $q$ are of course the projections of $X \times Y$ onto their components, and even more crucially, where any pair of other mappings $T \xrightarrow[]{y} Y$ and $T \xrightarrow[]{x} X$ must factor through the product object $X \times Y$ in a unique way. 
The mappings $y = q \circ r$ and $x = p \circ r$ by commutativity (where $q \circ r$ means apply $r$ first, and then apply $q$). In other words, the ``object" $X \times Y$ is not defined intrinsically by its parts (namely ordered pairs), but rather ``extrinsically" by the mappings that lead into it and out of it. Given additional morphisms $Y \xrightarrow[]{g} Z$ and $X \xrightarrow[]{f}$ that map to a common co-domain object $Z$, the object $X \times Y$ can be viewed as a {\em pullback}, which is a special case of a far more general construction in category theory called a {\em limit}. 

\begin{figure}[h]
\begin{center}
\begin{tikzcd}
  T
  \arrow[drr, bend left, "x"]
  \arrow[ddr, bend right, "y"]
  \arrow[dr, dotted, "r" description] & & \\
    & X  \times Y \arrow[r, "p"] \arrow[d, "q"]
      & X \arrow[d, "f"] \\
& Y \arrow[r, "g"] &Z
\end{tikzcd}
\end{center} 
\caption{Diagrams form an integral part of category theory. The abstract category of diagrams is a  {\em functor} $F$ that maps from some indexing category of ``abstract" diagrams ${\cal J}$ into the category ${\cal C}$. In this example, the abstract category ${\cal J}$ can be defined as $\bullet \leftarrow \bullet \rightarrow \bullet$, an example of a universal construction called a ``pullback". Diagrams formally can be defined to have {\em limits} and {\em colimits}, which are special types of {\em cones} or {\em cocones}. In this example shown, the objects $T$ and $X \times Y$ are both cones above the diagram $\bullet \leftarrow \bullet \rightarrow \bullet$, but only $X \times Y$ is a limit of this diagram because every other cone must factor through it, as shown. Limits and co-limits play an integral role in Universal Causality.} 
\label{diagrams}
\end{figure}
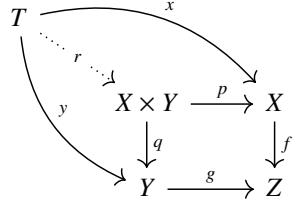 

As Figure~\ref{diagrams} shows, diagrams are not just a notational convenience in category theory. Rather, they are fundamental and a ``diagram" $F: {\cal J} \rightarrow {\cal C}$ is a {\em functor} $F$ that maps from some indexing category of ``abstract" diagrams ${\cal J}$ into the category ${\cal C}$ that represents a collection of mappings, one set that maps every object $X$ in the indexing diagram category ${\cal J}$ into the object $F(X)$ in the base category ${\cal C}$, and another set that maps every morphism $f: X \rightarrow Y$ in ${\cal J}$ into the corresponding morphism $F(f): F(X) \rightarrow F(Y)$. 

At its heart,  category theory is a science of analogy, where strict mathematical equality is replaced by isomorphisms, and looser notions like adjunctions and Kan extensions \cite{riehl2017category}. It defines objects not through their internal structure, as set theory does, but rather through their interaction with other objects. The membership $\in$ relation is replaced by an arrow $\rightarrow$, for example selecting an element of a set $X$ can be viewed as an arrow from a unit object ${\bf 1}$ to an object $X$ that is labeled with the element being selected (see Figure~\ref{covid-category}).  This notion can be immediately seen as a way of modeling causal intervention in Pearl's do-calculus framework \cite{pearl-book}.  Category theory has in recent years become an increasingly useful applied framework. \citet{fong2018seven} provide an excellent book-length treatment of the manifold ways in which category theory, specifically symmetric monoidal categories, can be used to reason about many complex problems in engineering. Within machine learning, one of the most successful and most cited  unsupervised dimensionality reduction methods is called UMAP, which originated entirely from using category theory \cite{mcinnes2018uniform}. A longstanding impossibility theorem on clustering was resolved successfully by treating clustering as a functor within category theory \cite{DBLP:journals/focm/CarlssonM13}. Finally, category theory is not only mathematically rigorous, but it also has an appealing diagrammatic structure that has found to be useful in a variety of applications \cite{fong2018seven,Baez_2010}. Category theory has also had a rich history in the design of functional programming languages, by treating programs as morphisms and types as objects \cite{DBLP:conf/esop/Reynolds09}. Some further examples of categories are illustrated below, which will be useful later in this paper. 

\begin{itemize} 

\item {\bf DAG:} The category {\bf DAG} has directed acyclic graphs (DAGs) as its objects, and morphisms correspond to paths (ignoring the directionality of the edges) from variable $X$ to $Y$. The unit morphism is the trivial path from $X$ to itself, and the composition of any two paths is also a path. The composition of paths is clearly associative.  Note that we can ``enrich" this category by selecting the set of morphisms {\bf Hom}$_{DAG}(X,Y)$ from another category ${\cal V}$, giving us the category of ${\cal V}$-enriched DAGs. The standard {\bf DAG} category can be viewed as a {\bf Set}-enriched category, if we simply define a set of morphisms from object $X$ to $Y$. 

\item {\bf ATop:} The category {\bf ATop} of all finite Alexandroff topological spaces \cite{alexandroff:1937} is defined as one where objects are finite topological spaces $(X, {\cal O}_X)$, where $X$ is a finite set, and ${\cal O}_X$ is a collection of ``open" subsets of $X$ closed under arbitrary union and intersection. The morphisms are all continuous functions from any object $f: (X, {\cal O}_X) \rightarrow (Y, {\cal O}_Y)$. Here, continuity is meant to imply that the pre-image $f^{-1}(B)$ of any open set $B \in {\cal O}_Y$ is also an open set in ${\cal O}_X$. Remarkably, as shown in our previous work \cite{sm:homotopy}, finite Alexandroff spaces are {\em directional}, and faithfully capture any causal graphical model, from a regular DAG to a hyper-edge DAG model \cite{hedge}.

\item {\bf Poset:} The category {\bf Poset} has partially ordered sets as its objects and order-preserving functions as its morphisms. That is, if $(X, \leq)$ and $(Y, \leq)$ are two poset objects in the category {\bf Poset}, the order-preserving morphism $f: (X, \leq) \rightarrow (Y, \leq)$ is defined by the property $x1 \leq x2, x1, x2 \in X$ implies that $f(x_1) \leq f(x_2)$ holds in $Y$. Interestingly, the category of all causal DAG  models can be faithfully embedded also in {\bf Poset}. 

\item {\bf Symmetric Monoidal Categories:} Many of the applications of category theory involve the use of symmetric monoidal categories, where objects can be composed in parallel or serial to form ``flowchart" like diagrams representing complex systems (see the book by \citet{fong2018seven}). Fong's unpublished 2012 MS thesis at the University of Oxford showed that Bayesian networks \cite{pearl:bnets-book} can be represented as symmetric monoidal categories fairly easily by defining the conditional probabilities of each variable $X$ in terms of its parents according to the DAG structure.

\item {\bf FinStat:} In the previous examples, we did not use a probability space, which we show how to do now. \citet{entropy-as-a-functor} proposed the category {\bf FinStat}, where objects are defined as $(X, p)$ where $X$ is a finite set and $p$ is probability distribution on $X$. A morphism in {\bf FinStat} is denoted by $(f,s): (X, p) \rightarrow (Y, r)$, and is comprised of a measure-preserving mapping $f: X \rightarrow Y$ together with a probability distribution $x \rightarrow s_{xy}$ on X for each element of $Y$ such that $s_{xy} =0 $ unless $f(x) = y$. We can visualize this morphism as a deterministic measurement from the finite set $X$ to the states $Y$ of the measurement apparatus. The measure-preserving property of $f$ implies that $r_y = \sum_{y: f(x) = y} p_x$. The $s$ component of the morphism represents a ``hypothesis" that the original system $X$ is in state $x$ with probability $s_{xy}$ when the measurement state is $y$. Clearly, {\bf FinStat} can be used to represent each edge in a standard Bayesian network causal model of a directed edge from node $X \rightarrow Y$, where $p(X)$ is the marginal distribution of $X$, and the marginal distribution $P(Y)$ is defined by the conditional probability $P(Y | X)$ and the marginal distribution $P(X)$. 

\item {\bf Meas:} The concept of  {\em information fields}, proposed by \citet{witsenhausen:1975}, can be used to model causal inference in a graph-independent way using causal information fields \cite{heymann:if}. To capture causal information fields, we can define a category of measurable spaces {\bf Meas}, where each object is a measurable space, defined by the pair $(X, {\cal B}_X)$, where $X$ is some arbitrary set, and ${\cal B}_X$ is a $\sigma$-algebra over $X$. Morphisms correspond to measurable functions from $f: (X, {\cal B} \rightarrow (Y, {\cal B}_Y)$. 
\end{itemize}

\subsection{The Theory of Sheaves} 

Many ideas in category theory, such as the generalization of sets into toposes \cite{goldblatt:topos}, came about largely due to the attempt to build a mathematical theory of ``data structures", namely how to take ``local" pieces of information and fuse them together to build a coherent ``global" object. A classic example is the construction of sheaves in topology, which defines a continuous function $f: X \rightarrow Y$ from a topological space $(X, {\cal T}_X)$ to another space $(Y, {\cal T}_Y)$ as one that maps open sets to open sets, namely if $B \subseteq {\cal T}_Y$ is an open set in $Y$, then the pre-image of $f$ is open in $X$, namely $f(B)^{-1} \in {\cal T}_X$. A challenge in the theory of sheaves is how to ``collate" or assemble pieces of $f$ from components $f_i$ defined  over subsets of a topological space. 

We view both Bayesian networks and SCMs as examples of ``sheaves", namely data structures on some category that are ``collatable" in that local representations can be fused to form globally consistent models. The theory of sheaves \cite{maclane:sheaves} provides an exceptionally rich set of ways of building complex models from simpler pieces.  A Bayesian network defines a joint probability distribution 
 \[ P(X_1, \ldots, X_n) = \prod_{i=1}^n P(X_i | \mbox{Pa}(X_i)], \]  where 
$\mbox{Pa}(X_i) \subset \{X_1, \ldots, X_n \} \setminus {X_i}$ represents a subset of variables (not including the variable itself). Each term in this product is a ``sheaf", a local data structure, and the entire distribution is comprised of the assembly of sheaves collated together in a consistent manner. \citet{fong:ms} shows how Bayesian network models can be constructed using symmetric monoidal categories, where the tensor product operation is used to combine multiple variables into a ``tensored" variable that then probabilistically maps into an output variable. 

One goal in sheaf theory is to require that functions $f_i: U_i \rightarrow \mathbb{R}$, where  the collection of all sets $U_i$ represents an ```open cover" of $U$, be such that there is at most one continuous function $f: X \rightarrow \mathbb{R}$, which when restricted to $f |_{U_i}$ gives us the local pieces. Note the strong resemblance of sheaf theory to Bayesian networks and SCMs, where the global distribution or causal model is being ``assembled" from local conditional probability distributions or local causal laws defining the local functions $f_i$ in an SCM. This connection to sheaves shows how to generalize measure-theoretic generalizations of Bayesian networks, such as causal information fields \cite{heymann:if}, using a sheaf-theoretic view. 

\section{Universal Causal Model}

We now state more formally the Universal Causal Model (UCM) that underlies our UC framework, which can be viewed as a broad generalization of graph-based paradigms such as Structural Causal Models (SCM) \cite{pearl-book,spirtes:book} or structural equation models (SEMs) \cite{sem:book}to arbitrary causal categories.  Our goal is develop a representation that is rich enough to cover the gamut of approaches illustrated in Table~\ref{causal-cats}. Accordingly, our formalism is not restricted to graph-based or logical or probabilistic representations, such as used in potential outcomes \cite{rubin-book} or Pearl's do-calculus \cite{pearl-book}, or actual causality \cite{halpern:ac} or Lewis' counterfactuals \cite{DBLP:journals/jphil/Lewis73}. The UC framework rather is based entirely on the structures and concepts that arise out of category theory. 

\begin{figure}[t]
\centering
\begin{minipage}{0.7\textwidth}
\centering
\hfill
\includegraphics[scale=0.45]{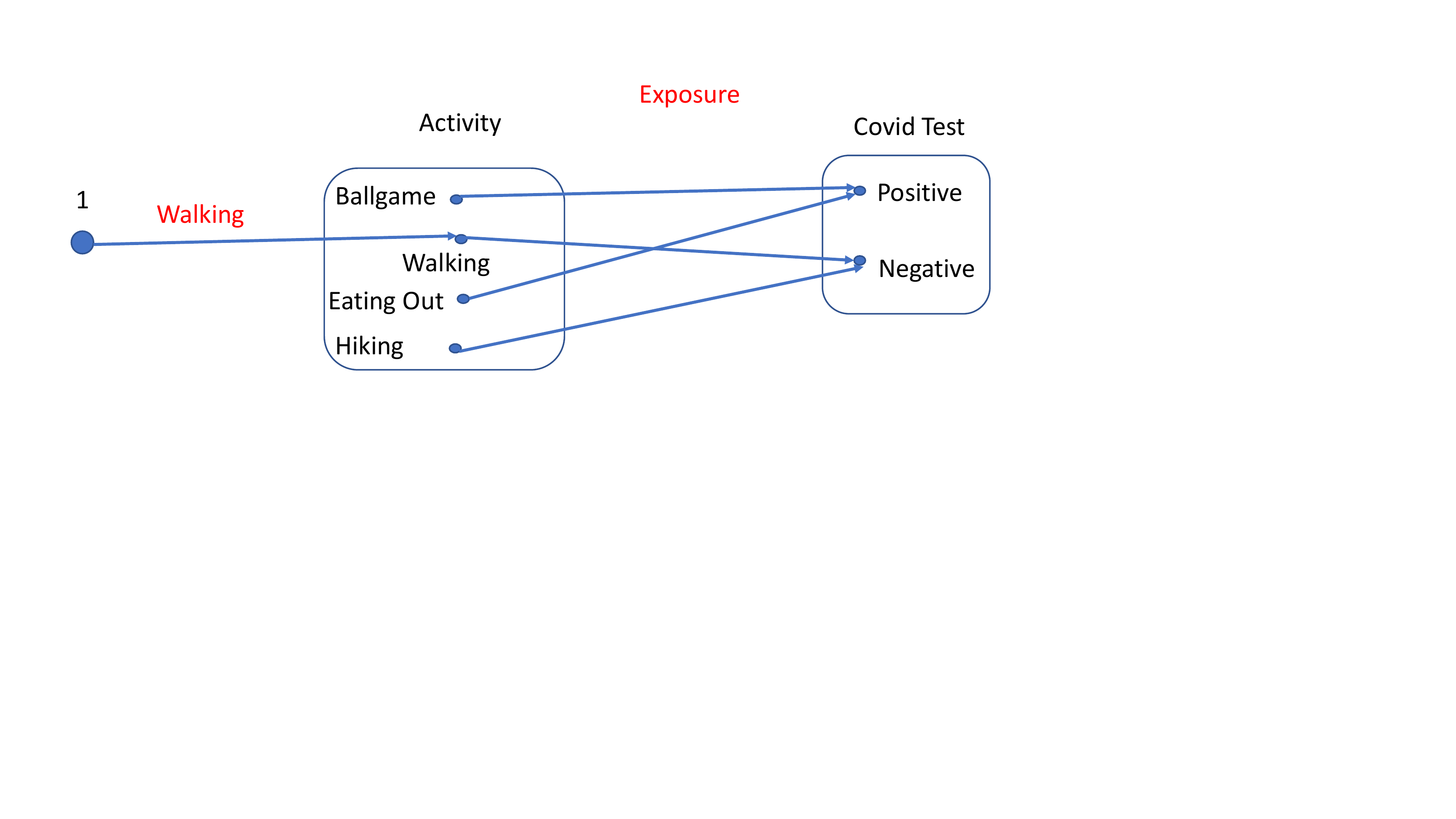}
\end{minipage} \vskip -1.5in
\caption{A simple category ${\cal C}$ for studying causal effects of exposure to getting Covid-19 infections. The morphism {\bf Exposure } maps a finite set object {\bf Activity} to the (Boolean) object  {\bf Covid Test}, where the morphism could represent any arbitrary  measure-theoretic \cite{witsenhausen:1975, heymann:if}, probabilistic \cite{pearl-book}, or topological relationship \cite{sm:homotopy}.  In category theory, an ``element" of a set (such as {\bf Walking}) is formally denoted by a morphism from the unit object {\bf 1} (which has one object and one identity morphism) to the set that is defined by the label it maps to. In UC, causal interventions are viewed as morphisms into an object, such as determining the causal effect of {\bf Activity} on {\bf Covid Test} by intervening on the {\bf Activity} variable by setting it to {\bf Walking}. UC applies to morphisms over any category of objects, whether they are discrete functions over finite sets, continuous functions over topological spaces \cite{sm:homotopy}, measurable functions over measurable spaces \cite{heymann:if},  or stochastic maps \cite{Baez_2010}. UC can also represent edge-interventions \cite{janzing}, where a variable being intervened on is assumed to take on any value with some pre-defined marginal distribution. For example, instead of intervening on {\bf Activity} by setting it to {\bf Walking}, we can instead intervene on {\bf Activity} by choosing some distribution over the set of activities.}  
\label{covid-category}
\end{figure}

\begin{definition}
A {\bf universal causal model} (UCM) is defined as a tuple $\langle {\cal C}, {\cal X}, {\cal I}, {\cal O}, {\cal E} \rangle$ where each of the components is specified as follows: 

\begin{enumerate} 
\item ${\cal C}$ is a category of causal objects that interact with each other, and their patterns of interactions are captured by a set of morphisms {\bf Hom}$_{\cal C}(X,Y)$ between object $X$ and $Y$. Categories where the {\bf Hom} morphisms can be defined as a set are referred to as {\em locally small} categories, which are the ones of primary interest to us in this paper. In particular, a {\cal V}-enriched category ${\cal C}$ is one where the {\bf Hom} morphisms are defined over the category {\cal V}, which allows capturing additional structure that might exist in the set of morphisms. 

\item ${\cal X}$ is a set of {\em construction} objects (see Figure~\ref{operads}) that allow constructing complex causal models from elementary parts. Examples include monoidal tensor product of two categories ${\cal M} \otimes {\cal M}$, Galois extensions and profunctors between two partially ordered sets to represent resource models,  co-limits and decorated cospans to represent electric circuits, and so on. 

\item ${\cal I}$ is an intervention category of intervention objects, which map from some category ${\cal E}$ of experimental designs into the causal category ${\cal C}$ to implement an experimental design. Interventions can be simple, such as setting the value of a variable to a specific value (see Figure~\ref{covid-category}), in which case ${\cal I}$ is a comma category, or they can be more complex, such as setting the value of a variable to some arbitrary marginal distribution, such as the edge-intervention model proposed by \citet{janzing}.  There is a large literature on treatment planning in causal inference, and many of the state of the art applications of causal inference, such as bipartite experiments \cite{zigler2018bipartite} use sophisticated experimental designs. Any of these experimental designs can be accommodated in our functorial intervention framework. 

\item ${\cal O}$ is a functor category of observation objects, which provide a (partial, perhaps noisy) view of causal objects in ${\cal C}$. For example, an object in the covariant functor category of presheafs {\cal Hom}$_{\cal C}(X, -)$ of morphisms out of $X$ can be viewed as a measurement of the ``state" of $X$.  More general examples include {\em bisimulation morphisms} used in the literature in reinforcement learning and software systems \cite{sm:udm}. \citet{adams} propose modeling observation as an order-preserving function $\Phi$ from a system ${\cal S}$ modeled by a lattice to an observation structure ${\cal O}$, which is also a lattice. They define the composition of subsystems into a larger system as a {\em join} $S_1 \vee S_2$, where a system exhibits a generative effect if $\Phi(S_1 \vee S_2) \neq \Phi(S_1) \vee \Phi(S_2)$. We can interpret $\Phi$ in our UC framework as an observation morphism that is not join preserving. 

\item ${\cal E}$ is a functor category of evaluation objects, which map a given causal model into an evaluation category for the purposes of evaluating the effects of an intervention. For an example, for a network economy \cite{nagurney:vibook}, the evaluation functor {\cal E} is defined as a mapping from the causal network economy model to the category of real-valued vector spaces defined by the vector field $F$ of a variational inequality (VI) $\langle F(x^*), (x - x^*) \rangle \geq 0, \forall x \in$. Here, the vector field $F: {\cal C} \rightarrow \mathbb{R}^n$ is a functor from the causal category ${\cal C}$ of network economy models (see \cite{sm:udm}) to the category of $n$-dimensional Euclidean space. Solving a VI means finding equilibrium flows $x^*$ on the network that represent stable patterns of trade. For example, due to the recent war in Ukraine, global supply chains from grain to natural gas and oil have been disrupted, and this intervention requires the global economic system to find a new equilibrium. The problem of studying causal interventions in network economies was recently studied by us in a previous paper \cite{sm:causal-network-econ}, and we refer the reader to that paper for more details. In a causal DAG model, the evaluation functor maps the causal category into the category of probability spaces (i.e., the  rules of Pearl's do-calculus \cite{pearl-book} give conditions under which $P(Y | do(X)) = P(Y | X)$).  
\end{enumerate} 

\end{definition}

\section{Functors, Natural Transformations and the Yoneda Lemma} 

At the outset, we noted that the principal goal of our paper is to define causal inference in terms of an underlying universal property. To make this notion more precise requires introducing some more advanced ideas from category theory. The notion of a universal property in UC is based on functors, which generalize the more familiar notion of functions in set theory. Functors map from one category to another, preserving the underlying structure of morphisms. {\em Natural transformations} are mappings from one functor to another, which play a key role in the UC framework. Our goal in the subsequent section is to use this machinery to define universal representations of causal models in the next section. 

\subsection{Causal Isomorphism in UC} 

Category theory can be viewed as the ``science of analogy". Instead of asking the question whether two objects are ``equal", it instead poses the question of whether objects are {\em isomorphic}. This notion of isomorphism is fundamental to our UC framework. We want to know if under the influence of a particular treatment $X$, an outcome $Y$ has changed in some way. This causal inference question is equivalent to asking whether the object $Y'$ (under treatment) is isomorphic to the object $Y$ under control. In the potential outcomes framework \cite{rubin-book}, the notation used is $Y(0)$ (potential outcome under control) and $Y(1)$ (potential outcome under treatment). The phrase ``potential outcome" is used because a unit under treatment cannot be observed under control and vice versa. In Pearl's do-calculus, causal inference is posed as the equation whether $P(Y | do(X=0)) = P(Y | do(X=1))$, or more generally, whether $P(Y | do (X)) = P(Y | X)$, also known as causal identifiability. In UC, all causal inference questions are posed as involving isomorphism of objects. Later, we will see weaker notions than isomorphism, including adjunctions and Kan extensions, which are going to be more useful for applying UC to real world problems. 

\begin{definition}
Two objects $X$ and $Y$ in a category ${\cal C}$ are deemed {\bf isomorphic}, or $X \simeq Y$ if and only if there is an invertible morphism $f: X \rightarrow Y$, namely $f$ is both {\em left invertible} using a morphism $g: Y \rightarrow X$ so that $g \circ f = $ {\bf id}$_X$, and $f$ is {\em right invertible} using a morphism $h$ where $f \circ h = $ {\bf id}$_Y$. 
\end{definition}

\subsection{Covariant and Contravariant Functors} 

UC is based on constructing covariant and contravariant functorial representations of causal models.

\begin{definition} 
A {\bf covariant functor} $F: {\cal C} \rightarrow {\cal D}$ from category ${\cal C}$ to category ${\cal D}$ is defined as the following: 
\begin{itemize} 
    \item An object ${\cal F}X$ of the category ${\cal D}$ for each object $X$ in category ${\cal C}$.
    \item A morphism ${\cal F}f: {\cal F}X \rightarrow {\cal F}Y$ in category ${\cal D}$ for every morphism $f: X \rightarrow Y$ in category ${\cal C}$. 
   \item The preservation of identity and composition: ${\cal F} \ id_X = id_{{\cal F}X}$ and $({\cal F} g) ({\cal F} g) = {\cal F}(f g)$ for any composable morphisms $f: X \rightarrow Y, g: Y \rightarrow Z$. 
\end{itemize}
\end{definition} 

\begin{definition} 
A {\bf contravariant functor} $F: {\cal C} \rightarrow {\cal D}$ from category ${\cal C}$ to category ${\cal D}$ is defined exactly like the covariant functor, except all the mappings are reversed. In the contravariant functor ${\cal F}: C^{\mbox{op}} \rightarrow D$, every morphism $f: X \rightarrow Y$ is assigned the reverse morphism ${\cal F} f: {\cal F} Y \rightarrow {\cal F} X$ in category ${\cal D}$. 
\end{definition} 

We introduce the following functors that will prove of value below: 

\begin{itemize} 
\item For every object $X$ in a category ${\cal C}$, there exists a covariant functor ${\cal C}(X, -): {\cal C} \rightarrow {\bf Set}$ that assigns to each object $Z$ in ${\cal C}$ the set of morphisms ${\cal C}(X,Z)$, and to each morphism $f: Y \rightarrow Z$, the pushforward mapping $f_*: {\cal C}(X,Y) \rightarrow {\cal C}(X, Z)$. 

\item For every object $X$ in a category ${\cal C}$, there exists a contravariant functor ${\cal C}(-, X): {\cal C}^{\mbox{op}} \rightarrow {\bf Set}$ that assigns to each object $Z$ in ${\cal C}$ the set of morphisms ${\cal C}(X,Z)$, and to each morphism $f: Y \rightarrow Z$, the pullback mapping $f^*: {\cal C}(Z, X) \rightarrow {\cal C}(Y, X)$. Note how ``contravariance" implies the morphisms in the original category are reversed through the functorial mapping, whereas in covariance, the morphisms are not flipped. 
\end{itemize} 

\begin{definition} 
\label{fully-faithful} 
Let ${\cal F}: {\cal C} \rightarrow {\cal D}$ be a functor from category ${\cal C}$ to category ${\cal D}$. If for all objects $X$ and $Y$ in ${\cal C}$, the map ${\cal C}(X, Y) \rightarrow {\cal D}({\cal F}X, {\cal F} Y)$, denoted as $f \mapsto {\cal F}f$ is
\begin{itemize}
    \item injective, then the functor ${\cal F}$ is defined to be {\bf faithful}. 
    \item surjective, then the functor ${\cal F}$ is defined to be {\bf full}.  
    \item bijective, then the functor ${\cal F}$ is defined to be {\bf fully faithful}. 
\end{itemize}

\end{definition} 

Our goal is to construct fully faithful functorial embeddings of causal models, which gives us an embedding of causal models into the category of sets. 

\subsection{Natural Transformations} 

Functors map from one category (e.g, all DAGs) to another category (e.g., finite topological spaces). Given two functors $F, G: {\cal C} \rightarrow {\cal D}$, we want to similarly define a mapping from one functor to the other. 

\begin{definition}
Given two functors ${\cal F}, {\cal G}: {\cal C} \rightarrow {\cal D}$ that map from category ${\cal C}$ to category ${\cal D}$, a {\em natural transformation} $\eta: {\cal F} \rightarrow {\cal G}$ consists of a morphism $\eta_X: {\cal F}X \rightarrow {\cal G} X$ for each object $X$ in ${\cal C}$. Moreover, these morphisms should satisfy the following property, that is the diagram below should commute: 
\begin{center}
    \begin{tikzcd}
  {\cal F}X \arrow[r, "{\cal F} f"] \arrow[d, "\eta_X" red]
    & {\cal F}Y \arrow[d, "\eta_Y" red] \\
  {\cal G}X  \arrow[r,  "{\cal G} f" ]
& {\cal G}Y
\end{tikzcd}
\end{center}
\end{definition}

\begin{definition}
For any two functors ${\cal F}, {\cal G}: {\cal C} \rightarrow {\cal D}$, let $\mbox{Nat}({\cal F}, {\cal G})$ denote the natural transformations from ${\cal F}$ to ${\cal G}$. If $\eta_X: {\cal F}X \rightarrow {\cal G} X$ is an isomorphism for each $X$ in category ${\cal C}$, then the natural transformation $\eta$ is called a {\bf natural isomorphism} and ${\cal F}$ and ${\cal G}$ are naturally isomorphic, denoted as ${\cal F} \cong {\cal G}$. 
\end{definition}

\subsection{The Yoneda Lemma} 

The machinery of natural transformations between functors enables making concrete the central philosophy underlying category theory, which is construct representations of objects in terms of their interactions with other objects. Unlike set theory, where an object like a set is defined by listing its elements, in category theory objects have no explicit internal structure, but rather are defined through the morphisms that define their interactions with respect to other objects. The celebrated Yoneda lemma makes this philosophical statement more precise. 

\begin{theorem}
{\bf Yoneda Lemma:} For every object $X$ in category ${\cal C}$, and every contravariant functor ${\cal F}: {\cal C}^{\mbox{Op}} \rightarrow {\bf Set}$, the set of natural transformations from ${\cal C}(-, X)$ to ${\cal F}$ is isomorphic to ${\cal F} X$. 
\end{theorem}

That is, the natural transformations from ${\cal C}(-, X)$ to ${\cal F}$ serve to fully characterize the object ${\cal F} X$ up to isomorphism. In the special circumstance when the set-valued functor ${\cal F} = {\cal C}(-, Y)$, the Yoneda lemma asserts that $\mbox{Nat}({\cal C}(-, X), {\cal C}(-, Y) \cong {\cal C}(X, Y)$. In other words, a pair of objects are isomorphic $X \cong Y$ if and only if the corresponding contravariant functors are isomorphic, namely ${\cal C}(-, X) \cong {\cal C}(-, Y)$. 

Although a full proof of the Yoneda Lemma is outside the scope of this paper (see \cite{maclane:71,riehl2017category}), a profound implication for causal inference is that the Yoneda Lemma reveals the deep topological structure of causal inference. A causal intervention is essentially an object in the functor category of presheaves, i.e. all morphisms into the object: in simpler terms, to intervene on an object $X$, we must select a morphism {\em into} the object from some other object or category. In the simplest case, in Pearl's celebrated do-calculus \cite{pearl-book}, intervention on a variable means selecting a specific value. This simple form of intervention is modeled by a morphism from the terminal object {\bf 1} to the variable $X$ being intervened. As the terminal object {\bf 1} contains only 1 object, and 1 (identity) morphism, of necessity, any morphism from {\bf 1} to $X$ chooses a value for $X$ (if $X$ is defined as a set).

As mentioned previously, causal discovery is often carried out by interventions on groups of variables (or objects in a causal category).Many previous studies of causal discovery from interventions, including the {\em conservative} family of intervention targets \citep{DBLP:journals/jmlr/HauserB12}, path queries \citep{DBLP:conf/nips/BelloH18a}, and {\em separating systems} of finite sets or graphs  \citep{DBLP:conf/uai/Eberhardt08,DBLP:journals/jmlr/HauserB12,DBLP:conf/nips/KocaogluSB17,sepsets,MAOCHENG198415} can all be viewed functorially as maps from the category of finite topological spaces into the causal category that imposes a particular intervention topology. A very related notion is that of {\em separating systems} of finite sets as intervention targets  \citep{DBLP:conf/uai/Eberhardt08,DBLP:journals/jmlr/HauserB12,DBLP:conf/nips/KocaogluSB17}. or separating systems of graphs \citep{MAOCHENG198415,DBLP:journals/corr/hauser-arxiv}. \citet{DBLP:conf/nips/KocaogluSB17} used {\em antichains}, a partitioning of a poset into subsets of non-comparable elements. \citet{DBLP:conf/nips/BelloH18a} use path queries, which can be viewed as chains. Finally, \citet{prasad:aaai21} used leaf queries on tree structures, where none of the interior nodes can be intervened on. All of these cases naturally have a functorial interpretation as choosing some slice of a category. A much more general paradigm is to assume that an intervention is essentially a subobject classifier in a topos.

\subsection{Presheaf Representations} 

A very important class of representations that follow from the Yoneda lemma are {\em presheafs} ${\cal C}(-, X)$. Given any two categories ${\cal C}, {\cal D}$, we can always define the new category ${\cal D}^{\cal C}$, whose objects are functors ${\cal C} \rightarrow {\cal D}$, and whose morphisms are natural transformations. If we take ${\cal D} = {\bf Set}$, and consider the contravariant version ${{\bf Set}^{\cal C}}^{\mbox{op}}$, we obtain a category whose objects are presheafs. Presheafs have some very nice properties, which makes them a {\em topos} \citep{maclane:sheaves}. 

\subsection{Topoi} 

The category {\bf Set} have certain essential properties, such as admitting an initial and terminal element, having all finite limits and co-limits, and allowing exponentiation (function objects). These properties can be abstracted into a category called an {\em elementary topos} \cite{maclane:71,goldblatt:topos,Johnstone:592033}.  Topos theory becomes of central relevance to causal inference in the UC framework for two reasons. The UC framework formulates causal inference in the functor category of presheaves $\hat{{\cal C}}$, which form a topos. Further,  in each topos, there is an internal logic that can be used to model a fundamental type of counterfactual reasoning, such as the reasoning patterns studied by Lewis \cite{DBLP:journals/jphil/Lewis73}.  We will show below how to extend Pearl's do-calculus over graphs to the topos category of sheaves. 

 \section{Representability, Universal Properties, and Universal Elements} 

One of the most significant ideas in category theory, and possibly one of the hardest to appreciate initially, is the concept of a {\em universal property}, which is discussed in detail in Chapter 2 of the book by \citet{riehl2017category}. Our explanation follows her presentation. Universality at its core is also tied to the notion of {\em representability}. To say a property is universal is equivalent to saying it is representable through a so-called {\em universal element}. Our goal is to define the universal property that underlies causality, which forms the basis behind the UC framework. 

\begin{figure}[h]
\centering
\begin{minipage}{0.5\textwidth}
\centering
\includegraphics[scale=0.35]{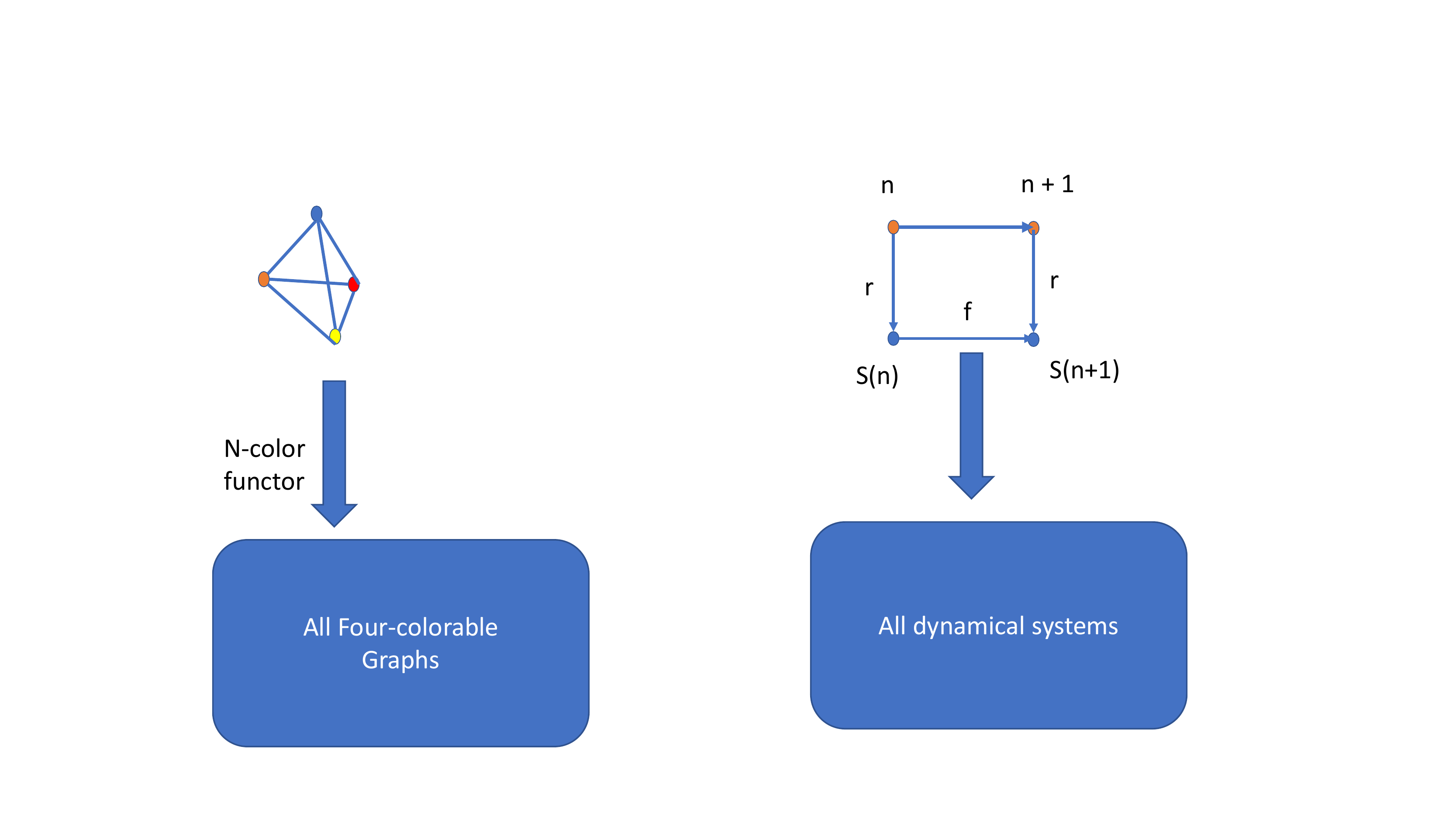}
\end{minipage}
\caption{To say a property is universal is to imply that it functorially representable by a {\em universal element}. The $N$-color functor maps any undirected graph $G$ from the category of graphs ${\cal G}$ to the category {\bf Set} representing its coloring such that no two adjacent vertices have the same color. The complete graph $K_N$ on $N$ vertices is the  graph with the smallest number of edges and vertices that can be colored with no less than $N$ colors. Consequently, it represents the universal element for the functor $N$-color that maps any graph to the set of colors that can be used to color the graph. Similarly, the natural numbers $\mathbb{N}$ is the {\em universal dynamical system} because every other dynamical system can be expressed by a morphism from it.} 
\label{universal-element}
\end{figure}

 The concept is explained in Figure~\ref{universal-element}. Consider the functor $N$-color: {\bf Graph} $\rightarrow {\bf Set}$ that maps from  category of all undirected graphs to the category of sets, such that $N$-color(G) is defined as the set of colors that can color the vertices of the graph such that no two adjacent vertices get the same color. Clearly, $K_N$, the complete graph on $N$ vertices is the {\em smallest} graph in terms of number of edges and vertices that can be colored with no fewer than $N$ colors. For example, for $N=4$, the famous Four-color theorem tells us that all planar graphs of any size can be colored with 4 colors. So, the complete graph $K_4$ shown is the universal element that {\em represents} the universal property of $4$-colorability. Similarly, let us consider the category of all discrete dynamical systems defined by an endomorphism $f: S \rightarrow S$, where $S$ is a finite set, and $S_0$ is some distinguished initial object. The natural numbers $\mathbb{N}$ under the successor function $s: \mathbb{N} \rightarrow \mathbb{N}$ defined as $s(n) = n+1$ is the {\em universal discrete dynamical system}, because given any discrete dynamical system, we can define a unique morphism $r: \mathbb{N} \rightarrow S$ such that the diagram on the right in Figure~\ref{universal-element} commutes, namely $S(n) = r(n)$.

\subsection{Diagrams as Functors, Limits and Co-Limits} 

The fundamental theorem of Universal Causality, defined in the next section, build on the concepts of co-limits and limits of causal diagrams. These are best understood first in a simpler setting. In many applications of mathematics involving convergence of algorithms, one typically asks for the limit of a sequence of iterations of the algorithm. In category theory, this concept is vastly generalized to the (co)limit of a diagram itself, where a diagram $J$ is itself viewed as a functor from the ``shape" of a diagram to the actual category. To make this somewhat abstract definition concrete, let us look at some simpler examples of universal properties, including co-products and quotients (which in set theory correspond to disjoint unions). Coproducts refer to the universal property of abstracting a group of elements into a larger one.

\begin{center}
\begin{tikzcd}
    & Z\arrow[r, "p"] \arrow[d, "q"]
      & X \arrow[d, "f"] \arrow[ddr, bend left, "h"]\\
& Y \arrow[r, "g"] \arrow[drr, bend right, "i"] &X \sqcup Y \arrow[dr, "r"]  \\ 
& & & R 
\end{tikzcd}
\end{center} 

In the commutative diagram above, the coproduct object $X \sqcup Y$ uniquely factorizes any mapping $h: X \rightarrow R$ and any mapping $i: Y \rightarrow R$, so that $h = r \circ f$, and furthermore $i = r \circ g$. Co-products are themselves special cases of the more general notion of co-limits. 

\subsubsection{Pullback Mappings} 

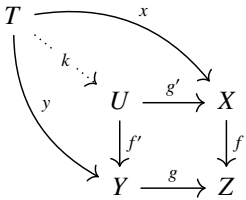
\begin{figure}[h]
\centering
\begin{tikzcd}
  T
  \arrow[drr, bend left, "x"]
  \arrow[ddr, bend right, "y"]
  \arrow[dr, dotted, "k" description] & & \\
    & 
    U\arrow[r, "g'"] \arrow[d, "f'"]
      & X \arrow[d, "f"] \\
& Y \arrow[r, "g"] &Z
\end{tikzcd}
\caption{Universal Property of pullback mappings.} 
\label{univpr}
\end{figure}

Figure~\ref{univpr}  illustrates the fundamental property of a {\em pullback}, which along with {\em pushforward}, is one of the core ideas in category theory. The pullback square with the objects $U,X, Y$ and $Z$ implies that the composite mappings $g \circ f'$ must equal $g' \circ f$. In this example, the morphisms $f$ and $g$ represent a {\em pullback} pair, as they share a common co-domain $Z$. The pair of morphisms $f', g'$ emanating from $U$ define a {\em cone}, because the pullback square ``commutes" appropriately. Thus, the pullback of the pair of morphisms $f, g$ with the common co-domain $Z$ is the pair of morphisms $f', g'$ with common domain $U$. Furthermore, to satisfy the universal property, given another pair of morphisms $x, y$ with common domain $T$, there must exist another morphism $k: T \rightarrow U$ that ``factorizes" $x, y$ appropriately, so that the composite morphisms $f' \ k = y$ and $g' \ k = x$. Here, $T$ and $U$ are referred to as {\em cones}, where $U$ is the limit of the set of all cones ``above" $Z$. If we reverse arrow directions appropriately, we get the corresponding notion of pushforward. So, in this example, the pair of morphisms $f', g'$ that share a common domain represent a pushforward pair.

Recall in the definition of a product that the object $X \times Y$ as well as $T$ played the role of cones. In fact, we can now see that the limit of the diagram functor ${\cal F}$ mapping the indexing category ${\cal D}: \bullet \rightarrow \bullet \leftarrow \bullet$ to the category ${\cal C}$ defined by the diagram $X \rightarrow Z \leftarrow Y$ is in fact the product element $X \times Y$, as every other cone $T$ must necessarily factor through it. Thus, $X \times Y$ represents the universal property of a product and it is also the limit of a diagram. Dually, for the co-product universal property, note that the co-product $X \sqcup Y$ is a co-limit because for any other cone $R$ ``under" the diagram must factor uniquely through it. Once again, $X \sqcup Y$ is an universal element representing a universal property.

\begin{definition}
Given a functor $F: {\cal J} \rightarrow {\cal C}$ from an indexing diagram category ${\cal J}$ to a category ${\cal C}$, an element $A$ from the set of natural transformations $N(A,F)$ is called a {\bf cone}. A {\bf limit} of the diagram $F: {\cal J} \rightarrow {\cal C}$ is a cone $\eta$ from an object lim $F$ to the diagram satisfying the universal property that for any other cone $\gamma$ from an object $B$ to the diagram, there is a unique morphism $h: B \rightarrow \mbox{lim} F$ so that $\gamma \bullet = \eta \bullet h$ for all objects $\bullet$ in ${\cal J}$. Dually, the {\bf co-limit} of the diagram $F: {\cal J} \rightarrow {\cal C}$ is a cone $\epsilon$ satisfying the universal property that for any other cone $\gamma$ from the diagram to the object $B$, there is a unique mapping $h: \mbox{colim} F \rightarrow B$ so that $\gamma \bullet = h \epsilon \bullet$ for all objects $\bullet$ in ${\cal J}$. 
\end{definition}

\section{Fundamental Theorems of Universal Causality} 

In this section, we state the Universal Causality Theorem and the Causal Representer Property. The first result establishes the universality of abstract causal diagrams, and second result shows the principal representer of causal information are presheaves.   These theoretical results are based directly on more basic results in category theory and the theory of sheaves.  Our primary contribution here is in recognizing their relevance and significance to causal inference, a connection that we have not seen before, to the best of our knowledge.  As we will see below, the universality property comes from formulating causal inference in the category of presheaves, where the universality of diagrams is a consequence of deep properties of the sheaf structure itself. In other words, the universality of causal inference is due entirely to the fact that it takes place in a presheaf, which is a topos \cite{maclane:sheaves}. This structure will lead us to understand the central result in the UC framework on the universality of causal diagrams. We emphasize that there is much work to be done here in elaborating the full consequences of these theorems, but a more detailed analysis is beyond the scope of this introductory paper. 

\subsection{The Fundamental Theorem of Universal Causality: Why Diagrams are Necessary} 

The above definitions play a major role in UC, because the Yoneda Lemma, perhaps the deepest result in category theory, defines the universal property of any object $X$ in a category ${\cal C}$ in terms of a universal element that also serves to represent it, namely the presheaf contravariant functor {\bf Hom}$_{\cal C}(-, X)$ of all morphisms into $X$. The presheaf is an element of the functor category {\bf Set}$^{{\cal C}^{op}}$. However, there is an even more elegant result that justifies the use of causal diagrams. The Yoneda Lemma asks us to pay attention to the contravariant functor of presheaves, namely {\cal Hom}$_{\cal C}(-, X)$, because the presheaf serves as a ``representer" of causal information.

We can now state a beautiful theoretical result -- the Universal Causality Theorem -- that serves to justify the term "Universal" in the Universal Causality framework. As this result is directly based on a result in the theory of sheaves, we first state that more basic result. 

\begin{theorem}
\label{presheaf-theorem}
In the functor category {\bf Set}$^{{\cal C}^{op}}$, every object $P$ is the colimit of a diagram of representable objects, in a canonical way. 
\end{theorem}

{\bf Proof:} This theorem is stated by \citet{maclane:sheaves} as Proposition 1 on page 41, Section 1.5 on colimits in functor categories in their book {\em Sheaves in Geometry and Logic}.  The theorem essentially states that given any functor $P$ in the category of presheaves (for example, the set of all causal influences on a variable $X$ in a causal model), there is a {\em canonical way} of constructing a small ``index" category ${\cal J}$ and a diagram (which recall is a functor) $A: {\cal J} \rightarrow {\cal C}$ of type ${\cal J}$ such that $P$ is isomorphic to the colimit of the composed diagram ${\cal J} \xrightarrow[]{A} {\cal C} \xrightarrow[]{y}$ {\bf Set}$^{{\cal C}^{op}}$, where the second mapping is the one given by the Yoneda embedding. The strategy behind proving this theorem is to show that the category of presheaves is Cartesian closed (see Figure~\ref{categories}), which have very attractive properties. A category is Cartesian closed if it has a terminal object {\bf 1}, it allows binary products of any two objects $X$ and $Y$, as well as exponentiation $Y^X$. For example, the category {\bf Set} is cartesian closed since clearly it has a terminal object {\bf 1} (the set with 1 object $ = \{ \bullet \}$, and one identity morphism), it allows taking Cartesian products, and it has function objects that are represented  by functions on sets. For the category of presheaves, showing the first two of these three properties is quite simple. The third, exponentiation, is where the proof is non-trivial, and the reader is referred to \cite{maclane:sheaves} for the last property. For the first two properties, note that the terminal object {\bf 1} in the presheaf category $\hat{{\cal C}}$ is simply the unique contravariant functor that maps the category ${\cal C}^{op}$ to the terminal object in {\bf Set}, that is every object $c$ in ${\cal C}$ is mapped to the one element set, and every morphism from $X$ to $Y$ in ${\cal C}$ is also mapped to the single morphism in {\bf 1}. For products, note that the product of two objects $P$ and $Q$ in the presheaf category $\hat{{\cal C}}$ is given by the pointwise product of $P(X)$ and $Q(X)$ over every object $X$ in ${\cal C}$. $\bullet$. 

\begin{theorem} 
\label{uct}
{\bf Universal Causal Theorem:} Given any universal causal model defined as a tuple ${\cal M} = \langle {\cal C}, {\cal X}, {\cal I}, {\cal O}, {\cal E} \rangle$, any causal inference in ${\cal M}$ can be represented as a co-limit of a diagram of representable objects in a unique way.
\end{theorem}

{\bf Proof:} The proof of this theorem builds directly on Theorem~\ref{presheaf-theorem}, which as we just stated is a basic result in the category theory of (pre)sheaves (see \cite{maclane:71}). This result states that {\em any} object $P$ in the functor category $\hat{{\cal C}} = $ {\bf Set}$^{{\cal C}^{op}}$ is representable uniquely as a co-limit of a diagram of representable objects. Of course, we already know from the Yoneda Lemma that the set of morphisms {\bf Hom}$_{\cal C}(-, X)$ into an object $X$ is an element of this functor category. As all causal inferences in UCM are carried out using presheaves (e.g, all queries that involve the analysis of paths that end in $X$ in a causal DAG, for example), it readily follows that given any contravariant functor $F: {\cal C}^{op}  \rightarrow $ {\bf Set}, there is a canonical way to construct a small index diagram $J$  such that the causal inference query $P$ is isomorphic to the co-limit of the diagram $F: {\cal J} \rightarrow {\cal C} \rightarrow$ {\bf Set}$^{{\cal C}^{op}}$ obtained exactly by the Yoneda Lemma defined above $\bullet$

{\bf This result, in short, can be viewed as the fundamental theorem underlying Universal Causality}. The central reason for its importance will become clearer below, but it asserts that Universal Causality has a universal element represented by a small index diagram. This somewhat abstract result can be viewed as justifying the foundational importance of abstract diagrams in causal reasoning. 

\subsection{Causal Reproducing Property} 

We now describe the second major result in UC, the Causal Reproducing Property (CRP), which states that any causal influence of an object $X$ upon any other object $Y$ can be represented as a natural transformation (a morphism) between two functor objections in the presheaf category $\hat{{\cal C}}$. The CRP is very akin to the idea of the reproducting property in kernel methods. Reproducing Kernel Hilbert Spaces (RKHS's)  transformed the study of machine learning, precisely because they are the unique subcategory in the category of all Hilbert spaces that have representers of evaluation defined by a kernel matrix $K(x,y)$ \cite{kernelbook}. The reproducing property in an RKHS is defined as $\langle K(x, -), K(-, y) \rangle = K(x,y)$. An analogous but far more general reproducing property holds in the UC framework, based on the Yoneda Lemma. 

\begin{theorem}
\label{crp}
{\bf Causal Reproducing Property:}  All causal influences between any two objects $X$ and $Y$ in a Universal Causal Model ${\cal M} = \langle {\cal C}, {\cal X}, {\cal I}, {\cal O}, {\cal E} \rangle$ can be defined from its presheaf functor objects, namely {\bf Hom}$_{\cal C}(X,Y) \simeq$ {\bf Nat}({\bf Hom}$_{\cal C}(-, X)$,{\bf Hom}$_{\cal C}(-, Y))$.  
\end{theorem}

{\bf Proof:} The proof of this theorem is a direct consequence of the Yoneda Lemma, which states that for every presheaf functor object $F$ in  $\hat{{\cal C}}$ of a category ${\cal C}$, {\bf Nat}({\bf Hom}$_{\cal C}(-, X), F) \simeq F X$. That is, elements of the set $F X$ are in $1-1$ bijections with natural transformations from the presheaf {\bf Hom}$_{\cal C}(-, X)$ to $F$. For the special case where the functor object $F = $ {\bf Hom}$_{\cal C}(-, Y)$, we get the result immediately that  {\bf Hom}$_{\cal C}(X,Y) \simeq$ {\bf Nat}({\bf Hom}$_{\cal C}(-, X)$,{\bf Hom}$_{\cal C}(-, Y))$. $\bullet$

The significance of the Causal Reproducing Property is that presheaves act as ``representers" of causal information, precisely analogous to how kernel matrices act as representers in an RKHS. 

\section{Kan Extensions of  Universal Causal Models} 

We now explore the use of Kan extensions, the single most powerful universal construction in category theory, from which every other concept can be defined. \citet{maclane:71} stated it boldly as ``Every concept is a Kan extension". We first introduce Kan extensions, and then provide a restatement of the Universal Causality Theorem (UCT) using Kan extensions.

\subsection{Kan Extension}

In this section, we briefly review the theory of Kan extensions \cite{maclane:71,riehl2017category}. It is well known in category theory that ultimately every concept, from products and co-products, limits and co-limits, and ultimately even the Yoneda embeddings, can be derived as special cases of the Kan extension \citep{maclane:71}. Kan extensions intuitively are a way to approximate a functor ${\cal F}$ so that its domain can be extended from a category ${\cal C}$ to another category  ${\cal D}$.  Because it may be impossible to make commutativity work in general, Kan extensions rely on natural transformations to make the extension be the best possible approximation to ${\cal F}$ along ${\cal K}$. 

\begin{definition}
A {\bf left Kan extension} of a functor $F: {\cal C} \rightarrow {\cal E}$ along another functor $K: {\cal C} \rightarrow {\cal D}$, is a functor $\mbox{Lan}_K F: {\cal D} \rightarrow {\cal E}$ with a natural transformation $\eta: F \rightarrow \mbox{Lan}_F \circ K$ such that for any other such pair $(G: {\cal D} \rightarrow {\cal E}, \gamma: F \rightarrow G K)$, $\gamma$ factors uniquely through $\eta$. In other words, there is a unique natural transformation $\alpha: \mbox{Lan}_F \implies G$. \\
%
\begin{center}
\begin{tikzcd}[row sep=2cm, column sep=2cm]
\mathcal{C}  \ar[dr, "K"', ""{name=K}]
            \ar[rr, "F", ""{name=F, below, near start, bend right}]&&
\mathcal{E}\\
& \mathcal{D}  \ar[ur, bend left, "\text{Lan}_KF", ""{name=Lan, below}]
                \ar[ur, bend right, "G"', ""{name=G}]
                
%
\arrow[Rightarrow, "\exists!", from=Lan, to=G]
\arrow[Rightarrow, from=F, to=K, "\eta"]
\end{tikzcd}
\end{center}
\end{definition}

A {\bf right Kan extension} can be defined similarly. 

\subsection{Universal Causality Theorem using Kan Extension} 

We now reformulate the Universal Causality Theorem (Theorem~\ref{uct}) using Kan extensions.

\begin{theorem} 
\label{uct-kan}
Given any universal causal model defined as a tuple ${\cal M} = \langle {\cal C}, {\cal X}, {\cal I}, {\cal O}, {\cal E} \rangle$, any causal inference in ${\cal M}$ can be represented as a left Kan extension of a diagram of representable objects in a unique way.
\end{theorem}

{\bf Proof:} Recall from the proof of the UCT (Theorem~\ref{uct}) that {\em any} object $P$ in the functor category $\hat{{\cal C}} = $ {\bf Set}$^{{\cal C}^{op}}$ is representable uniquely as a co-limit of a diagram of representable objects. As all causal inferences in UCM are carried out using presheaves (e.g, all queries that involve the analysis of paths that end in $X$ in a causal DAG, for example), given any contravariant functor $F: {\cal C}^{op}  \rightarrow $ {\bf Set}, there is a canonical way to construct a small index diagram $J$  such that the causal inference query $P$ is isomorphic to the co-limit of the diagram $F: {\cal J} \rightarrow {\cal C} \rightarrow$ {\bf Set}$^{{\cal C}^{op}}$ obtained exactly by the Yoneda Lemma. Now, the main step in the reformulation of the UCT using the Kan extension is to show that the co-limit of diagram $F$ in the original proof of UCT can be itself constructed as a Kan extension. This follows from a more general result that co-limits themselves are just left Kan extension of a functor $F: {\cal C} \rightarrow {\cal E}$ along the functor $L: {\cal C} \rightarrow$ {\bf 1}, where {\bf 1} is the terminal object in category ${\cal C}$. In this application, the functor being extended $F: {\cal J} \rightarrow {\cal C}$ is the diagram functor, as shown below. 

\begin{center}
\begin{tikzcd}[row sep=2cm, column sep=2cm]
\mathcal{J}  \ar[dr, "K"', ""{name=K}]
            \ar[rr, "F", ""{name=J, below, near start, bend right}]&&
\mathcal{C}\\
& {\bf 1}    \ar[ur, bend left, "\text{Lan}_KF", ""{name=Lan, below}]
                \ar[ur, bend right, "G"', ""{name=G}]
%
\arrow[Rightarrow, "\exists!", from=Lan, to=G]
\arrow[Rightarrow, from=J, to=K, "\eta"]
\end{tikzcd}
\end{center}

Note that the left Kan extension $\mbox{Lan}_K F:$ {\bf 1} $\rightarrow {\cal C}$ essentially has to select a object $c$ of ${\cal C}$, as {\bf 1} contains only one object and one identity mapping. Thus, composing $\mbox{Lan}_K F:$ {\bf 1}  $\rightarrow {\cal C}$ with the functor $K: {\cal J} \rightarrow$ {\bf 1} induces a constant functor $\Delta c$ from the diagram category ${\cal J}$ to the actual causal model category ${\cal C}$, where every object of ${\cal J}$ is mapped to object $c$, and every morphism $f: X \rightarrow Y$ in ${\cal J}$ is mapped to the unit morphism $1_c$ of $c$. The universality of the left Kan extension states that {\em every} other functor $G$ mapping the unit category {\bf 1} to ${\cal C}$ must factor through the left Kan extension in a unique way. Further, the natural transformation $\eta$ now maps every object and every morphism in ${\cal F}$ to the constant functor $\Delta_c$. Thus, the constant functor becomes a co-cone under the diagram functor, and universality guarantees that this co-cone is indeed a co-limit. $\bullet$ 

\subsection{Modeling Unobservable Confouders with Kan Extensions} 

We now address a key limitation of using the Yoneda embedding of presheaf functors to do causal inference. The Yoneda Lemma, as applied to causal inference, essentially states that presheaf functors capture everything that we need to know about an object $X$ to characterize it up to isomorphism. In other words, if $X$ is an outcome variable in a causal model, all influences on $X$ from other variables $Y$ intuitively do capture everything we need to know to do causal reasoning. Unfortunately, the reality is that unobservable confounders make it impossible in practice to use the desired presheaf functor {\bf Hom}$_{\cal C}(-, X)$. Thus, the key idea in UC to do causal inference with unobservable confounders is to use the Kan extension to {\em approximate the presheaf functor object} {\bf Hom}$_{\cal C}(-, X)$ along the {\em observable presheaf functor object} {\bf Hom}$_{\cal D}(-,X)$, where ${\cal D}$ is the subcategory  of ``observable" objects in the full category ${\cal C}$. The Kan extension shows us how to formulate this problem in a very general way. Of course, a full discussion of Universal Causal Models with confounders is beyond the scope of this paper, and will be the topic of a subsequent paper. 

\section{Hierarchical Abstraction of Universal Causal Models} 

One of the major strengths of the category theoretic framework is that it lets us build compositionally structured models in an exceptionally diverse and interesting set of ways \cite{fong2018seven} (see Table~\ref{operads}).  In this section, we show how to represent a universe of causal models, where each object represents an entire causal model, not a single causal variable as we have done above. This abstraction of entire causal models as objects reflects a systems-level perspective, which has shaped many areas of science and engineering. We think of the climate change problem as involving the interaction of many subsystems, including weather, the geology of the Earth, man-made effects etc., each of which can in turn be decomposed into many other subsystems (e.g, pollution to the burning of natural gas, emissions from vehicles etc.). 

\subsection{Causal Models as Objects} 

We define a new category ${\cal C}_{UCM}$ of all universal causal models, where each object $X$ is a universal causal model, and the morphisms map one UCM object into another. As in the previous discussion, the construction of this category can be done in many ways, where objects can be a DAG \cite{pearl-book} or one of its many variants \cite{hedge,mdag} or a causal information field \cite{heymann:if}. 

\begin{definition} 
The category of all universal causal models ${\cal C}_{UCM}$ is defined by a set of objects {\bf Obj}(${\cal C}_{UCM}$), where each object ${\cal M}$ is a UCM, and a set of morphisms $f: {\cal M}_1 \rightarrow {\cal M}_2$ from UCM object ${\cal M}_1$ to UCM object ${\cal M}_2$ that satisfy all the properties of being a category (i.e., there is an identity morphism for each UCM object ${\cal M}$, morphisms are composable into new morphisms, and satisfy the associative property. 
\end{definition} 

\subsection{Universal Categories of Causal Models} 

Depending on the types of UCM objects being considered and the type of morphisms, many examples of this construction can be given, some of which are listed below. 

\begin{figure}[h]
\centering
\begin{minipage}{0.5\textwidth}
\centering
\includegraphics[scale=0.35]{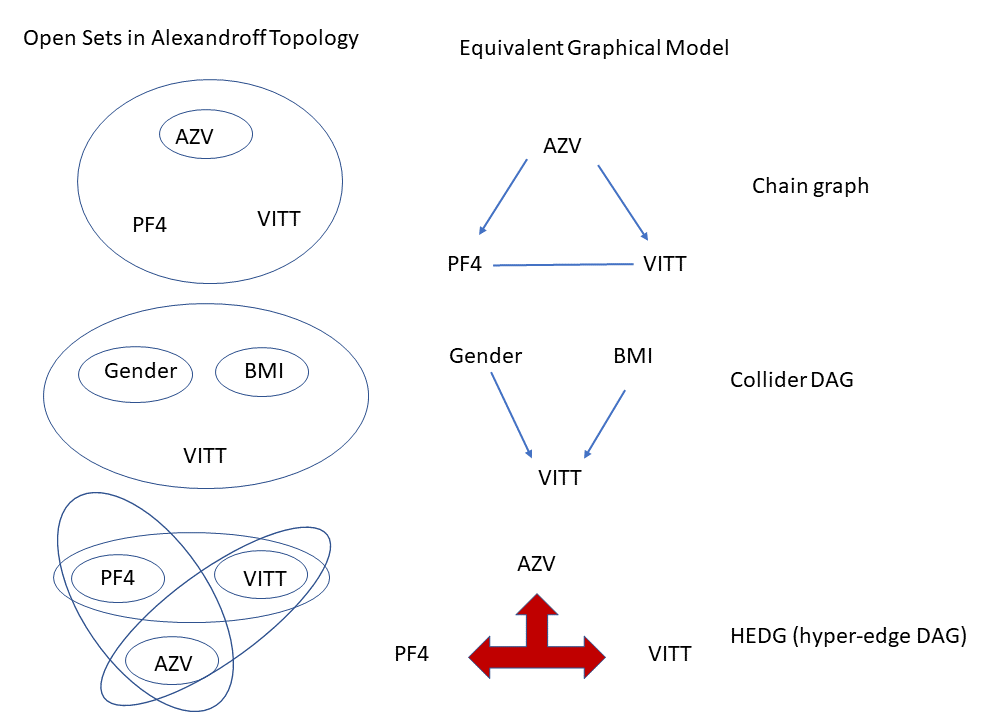}
\end{minipage}
\caption{Many causal graphical models, incuding DAGs, mDAGs, and HEDG models can all be represented as finite Alexandroff topological spaces, where every variable $X$ in the graphical model is associated with the open set of all its descendants (or the closed set of all its ancestors). The Alexandroff topological space is simply the topological space defined by all presheaf functor objects over all variables $X$. The models here illustrated come from a study of the side-efects of treatment of patients for the COVID pandemic using vaccines \citep{doi:10.1056/NEJMoa2104840,doi:10.1056/NEJMoa2104882}, where {\bf AZV} represents the adminstration of the AstraZeneca vaccine, {\bf PF4} denotes heparin-induced platelet factor 4, {\bf Gender} is important as many patients who exhibited adverse effects to the Covid vaccine were disproportionately women, {\bf HIT} denotes Heparin, a blood thinner used to prevent blood clots, and {\bf VITT} denotes  vaccine-induced immune thrombotic thrombocytopenia.} 
\label{alexandroff}
\end{figure}

\begin{enumerate} 

\item Define ${\cal C}_{DAG}$ as the category of all causal DAG models, where each object is a DAG model $(V, E, P)$ specified by a set of vertices, a set of edges $E$ and a probability distribution $P$ that factors according to the structure of the DAG (i.e., every conditional independence in the distribution $P$ is satisfied in the DAG, and vice versa). Morphisms are defined as causal interventions, where a variable $X$ in a model ${\cal M}$ is intervened on, resulting in a submodel ${\cal M}_{do(X)}$. This type of construction is common to many studies of causal inference in the Pearl paradigm \cite{pearl:causalitybook}. 

\item Define the category ${\cal C}_{ATop}$ of all finite Alexandroff topological space causal models \cite{alexandroff:1937,sm:homotopy} as the category where the objects are any finite Alexandroff topological space (or simply, finite space) $(X, \mathcal{O}_X)$, with a finite set $X$ and a collection $\mathcal{O}_X$ of ``open" sets, namely subsets of $X$,  such that (i) $\emptyset$ (the empty set) and $X$ are in $\mathcal{O}_X$ (ii) Any arbitrary union of sets in $\mathcal{O}_X$ is in $\mathcal{O}_X$ (iii) Any arbitrary intersection of subsets in $\mathcal{O}_X$ is in $\mathcal{O}_X$ as well. The morphisms in ${\cal C}_A$ are all continuous functions from one finite space to another, that is {\cal Hom}$_{{\cal C}_A}(X,Y)$ is the set of all functions from space $X$ to $Y$ that are continuous, meaning for any open set $O \in \mathcal{O}_Y$, the pre-image $f^{-1}(O)$ is an open set in $\mathcal{O}_X$. \citet{sm:homotopy} showed that many variants of causal graphical models, including DAGs \cite{pearl:causalitybook}, mDAGs \cite{mdag}, HEDG models \cite{hedge} etc.  can be represented as finite Alexandroff spaces (see Figure~\ref{alexandroff}). 

\item Define the category ${\cal C}_{IM}$ of intrinsic models, where each model is an instance of Witsenhausen's information field decision model \cite{witsenhausen:1975}. \citet{sm:udm} analyzed the properties of the category of intrinsic models in detail, but we briefly summarize the main ideas here. \citet{witsenhausen:1975} proposed the intrinsic model as a collection of variables $\alpha \in A$, where each variable $\alpha$ defines a measurable space $(U_\alpha, {\cal B}_\alpha)$, where ${\cal B}_\alpha$ is a sigma algebra over $U_\alpha$. Each variable $\alpha$ chooses a value $u \in U_{\alpha}$ using a policy $\pi_\alpha: U \rightarrow  U_\alpha$, where $U = \prod_\beta U_\beta$  is the product decision space, and each $\pi_\alpha$ is a measurable function over $(I_\alpha, {\cal B}_\alpha)$, where ${\cal I}_\alpha \subset \bigotimes_{\beta \in A} {\cal B}_\beta$ is the {\em information field} for variable $\alpha$. Witsenhausen's information field model was recently applied to causal inference, and shown to generalize Pearl's rules of do-calculus for causal DAG models \cite{heymann:if}. 

\item Define the category ${\cal C}_{DS}$ of {\em discourse sheaves} \cite{discourse-sheaves}, which represent an application of the theory of sheaves to model opinion formation in social networks. A discourse sheaf is a graph, where each vertex represents a person, and each edge represents communication between two people. Associated with each vertex is a {\em vertex sheaf} vector space, which represents the person's opinions on a variety of subjects (e.g., climate change, gun control, taxes etc.). Associated with each edge is an {\em edge sheaf} vector space that models communication between two individuals. The communication from individual $i$ to $j$ is modeled as a linear transformation from the vertex sheaf vector space to the edge sheaf vector space. The problem is to discover how opinions form in such social networks using a dynamical system model. In the category ${\cal C}_{DS}$, each object is a discourse sheaf, and morphisms between discourse sheaves represent changes in social networks as people of different views enter and leave the network.

\end{enumerate} 

It is clear from these examples that a very varied and rich class of causal models can be represented as objects within a larger universal category of all causal models of that type in the category. \citet{sm:udm} defines other examples, such as the category of all Markov decision processes, all predictive state representations, and all network economic models \cite{nagurney:vibook}. The framework of Universal Causality can be applied to all these interesting examples. 

\section{Beyond Graphs: Exploiting Richer Forms of Causal Interaction} 

\begin{figure}[h]
\centering
\begin{minipage}{0.5\textwidth}
\centering
\hfill
\includegraphics[scale=1]{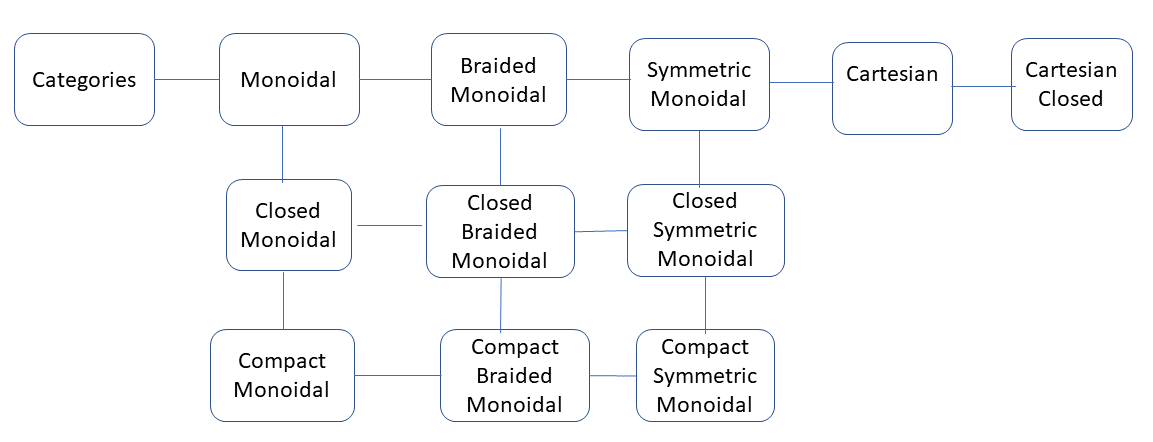}
\end{minipage}
\caption{Categories can be defined in many different ways. Listed here are a few broad types. Crucially, the UC framework shows that regardless of the original category where the model is defined, the presheaf functor over which Universal Causality is defined, forms a Cartesian closed category, which has very nice theoretical properties.} 
\label{categories}
\end{figure}

With the space of one paper, it is not practical to fully discuss the rich set of possibilities that emerge from combining category theory and causal inference. Figure~\ref{categories} lists a few of the many types of categories where causal inference can be explored in future work. Among these, symmetric monoidal categories have proven to be extremely useful in defining a rich set of applications \cite{fong2018seven}. In the proposed framework of Universal Causality, the most important abstraction to take away from this figure is that regardless of the original category over which the causal model is defined, the presheaf contravariant functor over which causal inference takes place is a Cartesian closed category. In other words, ``all roads lead to Rome" (i.e., Cartesian closed categories), as far as causal inference is concerned. This mapping from the original category to the category of Cartesian closed categories takes place principally through using the Yoneda Lemma, since presheaves {\cal Hom}${\cal C}(-, X)$ define a Cartesian closed category \cite{maclane:sheaves}. 

Finally, we want to illustrate how category theory provides a rich set of construction tools that facilitate modeling complex types of interactions among components in a causal network that go significantly beyond simple graphs that have been extensively studied in the causal inference literature to date. Many more examples of categories can be given, such as monoidal categories that require defining a tensor product operator $\otimes: {\cal C} \times {\cal C} \rightarrow {\cal C}$. Monoidal categories are incredibly useful in a variety of rich applications, as beautifully detailed in the book by \citet{fong2018seven}. Causal inference in many of these structures have not been studied to date, to the best of our knowledge. The UC framework applies to all of these interesting and diverse categorical representations. 

As Figure~\ref{operads} shows, category theory can be used in defining causal models over arbitrarily complex objects that interact in a number of non-directional ways, from electric circuits to signal flow graphs to complex engineered design structures, using some of the refined categorical structures shown in Figure~\ref{categories}. For example, braided categories are a type of monoidal category defined by a form of ``braiding" defined by the tensor product $X \otimes Y$ of two objects.  A detailed discussion of graphical languages for various types of monoidal categories is given in \cite{Selinger_2010}. The application of category theory to represent rich control structures is described in \cite{baez2015categories}. A full investigation of causal inference over models defined with these construction tools that category theory provides us is beyond the scope of this introductory paper. 

\begin{figure}[t]
\begin{center}
\begin{tabular}{|c |c | } \hline 
{\bf Construction Tool } & {\bf Application} \\ \hline 
Braiding and tensor products \cite{JOYAL1996164} & Control theory \cite{Baez_2010} \\ \hline 
 Galois Extensions and Profunctors \cite{fong2018seven}& Engineering design \cite{censi} \\ \hline 
Co-limits and Decorated cospans \cite{cospans}  & Electric Circuits \cite{fong2018seven} \\ \hline
Operads \cite{higher-operads} & Databases \cite{operads-db}  \\ \hline
Hypergraph categories \cite{fong2018seven} & Signal flow diagrams \\ \hline 
Bisimulation morphisms \cite{DBLP:conf/lics/JoyalNW93} & Concurrent systems \cite{bisim} \\ \hline
\end{tabular}
\end{center}
\caption{Category theory offers a rich palette of construction tools to build complex models.}
\label{operads}
\end{figure}

\section{Summary and Future Work} 

In this paper, we proposed  Universal Causality (UC), a category-theoretic framework that formalizes the universal property that underlies causal inference. The proposed UC framework formulates causal inference in the functor category of presheaves $\hat{{\cal C}}$ = {\bf Set}$^{{\cal C}^{op}}$. The {\bf Universal Causality Theorem} states that any causal inference, defined as an object in the contravariant functor category $\hat{{\cal C}}$ of presheaves, is representable as the co-limit of a small indexing diagram that serves as its universal element. The {\bf causal reproducing property} (CRP) states that all causal influences between two objects $X$ and $Y$ in a category ${\cal C}$ can be defined using its presheaf functor objects, namely {\bf Hom}$_{\cal C}(X,Y) \simeq$ {\bf Nat}({\bf Hom}$_{\cal C}(-, X)$,{\bf Hom}$_{\cal C}(-, Y))$. The CRP property is analogous to, but more general than, the reproducing property in Reproducing Kernel Hilbert Spaces that served as the foundation for kernel methods in machine learning. In effect, CRP states that any  presheaf functor object in $\hat{C}$ serves as a ``representer" of causal information in any category.  Causal interventions in UC are defined by contravariant functors that comprise of mappings into an object, while observations are viewed as covariant functors defined by the set of all morphisms {\cal Hom}$_{\cal C}(X, -)$ out of an object.

There are several limitations of the UC framework that need to be highlighted. It is a very broad framework that is couched in the language of category theory. In the metaphor used by Freeman Dyson in his article on ``Birds and Frogs" \cite{dyson:birds-and-frogs}, UC is a ``bird"-like theory that ``soars high" to capture a global view of causal inference, but misses a lot of details that ``frog"-like theories that work with particular representations or applications can capture.  However, to scale UC to practical problems, it is important to bring in both representation-specific and application-specific constraints that make causal inference feasible in practice. We also did not have space to fully articulate how Universal Causality applies to rich applications of symmetric monoidal categories that are described in \cite{fong2018seven}. In particular, profunctors are generalizations of relations in category theory, just as functors generalize functions in set theory, and UC extends nicely to profunctors. One interesting application of profunctors is to engineering design of complex objects \cite{censi}. The category of presheaves $\hat{{\cal C}}$ also forms a {\em topos}, a deep abstraction of the category of {\bf Set}, implying that many set-theoretic concepts in causal inference developed over the past 150 years may be significantly generalized to any topos. In particular, UC applies to Lewis style logical counterfactuals, for which a topos-theoretic framework has been proposed \cite{topos-counterfactual}. An in-depth description of categorial logic underlying topos theory is given in \citet{goldblatt:topos}, which may be applicable to models of actual causality \cite{halpern:ac}.


\begin{thebibliography}{69}
\providecommand{\natexlab}[1]{#1}
\providecommand{\url}[1]{\texttt{#1}}
\expandafter\ifx\csname urlstyle\endcsname\relax
  \providecommand{\doi}[1]{doi: #1}\else
  \providecommand{\doi}{doi: \begingroup \urlstyle{rm}\Url}\fi

\bibitem[Pearl(2009{\natexlab{a}})]{pearl:causalitybook}
Judea Pearl.
\newblock \emph{Causality: Models, Reasoning and Inference}.
\newblock Cambridge University Press, USA, 2nd edition, 2009{\natexlab{a}}.
\newblock ISBN 052189560X.

\bibitem[Westland(2019)]{sem:book}
J.~Christopher Westland.
\newblock \emph{Structural Equation Models - From Paths to Networks, Second
  Edition}, volume~22.
\newblock Springer, 2019.
\newblock ISBN 978-3-030-12507-3.
\newblock \doi{10.1007/978-3-030-12508-0}.
\newblock URL \url{https://doi.org/10.1007/978-3-030-12508-0}.

\bibitem[Heymann et~al.(2021)Heymann, de~Lara, and Chancelier]{heymann:if}
Benjamin Heymann, Michel de~Lara, and Jean-Philippe Chancelier.
\newblock Causal inference theory with information dependency models, 2021.
\newblock URL \url{https://arxiv.org/abs/2108.03099}.

\bibitem[Witsenhausen(1975)]{witsenhausen:1975}
H.~S. Witsenhausen.
\newblock The intrinsic model for discrete stochastic control: Some open
  problems.
\newblock In A.~Bensoussan and J.~L. Lions, editors, \emph{Control Theory,
  Numerical Methods and Computer Systems Modelling}, pages 322--335, Berlin,
  Heidelberg, 1975. Springer Berlin Heidelberg.
\newblock ISBN 978-3-642-46317-4.

\bibitem[Pearl(1989)]{pearl:bnets-book}
Judea Pearl.
\newblock \emph{Probabilistic reasoning in intelligent systems - networks of
  plausible inference}.
\newblock Morgan Kaufmann series in representation and reasoning. Morgan
  Kaufmann, 1989.

\bibitem[Mahadevan(2021{\natexlab{a}})]{sm:homotopy}
Sridhar Mahadevan.
\newblock Causal homotopy, 2021{\natexlab{a}}.
\newblock URL \url{https://arxiv.org/abs/2112.01847}.

\bibitem[MacLane and leke Moerdijk(1994)]{maclane:sheaves}
Saunders MacLane and leke Moerdijk.
\newblock \emph{Sheaves in Geometry and Logic: A First Introduction to Topos
  Theory}.
\newblock Springer, 1994.

\bibitem[MacLane(1971)]{maclane:71}
Saunders MacLane.
\newblock \emph{Categories for the Working Mathematician}.
\newblock Springer-Verlag, New York, 1971.
\newblock Graduate Texts in Mathematics, Vol. 5.

\bibitem[Sch\"{o}lkopf and Smola(2002)]{kernelbook}
B.~Sch\"{o}lkopf and A.~J. Smola.
\newblock \emph{Learning with Kernels: Support Vector Machines, Regularization,
  Optimization, and Beyond}.
\newblock MIT Press, 2002.

\bibitem[Plato(1971)]{plato}
Plato.
\newblock \emph{Timaeus and Critias}.
\newblock Penguin Classics, 1971.

\bibitem[Riehl(2017)]{riehl2017category}
E.~Riehl.
\newblock \emph{Category Theory in Context}.
\newblock Aurora: Dover Modern Math Originals. Dover Publications, 2017.
\newblock ISBN 9780486820804.
\newblock URL \url{https://books.google.com/books?id=6B9MDgAAQBAJ}.

\bibitem[Goldblatt(2006)]{goldblatt:topos}
Robert Goldblatt.
\newblock \emph{Topoi: The Categorial Analysis of Logic}.
\newblock Dover Press, 2006.

\bibitem[Johnstone(2002)]{Johnstone:592033}
Peter~T Johnstone.
\newblock \emph{{Sketches of an elephant: a Topos theory compendium}}.
\newblock Oxford logic guides. Oxford Univ. Press, New York, NY, 2002.
\newblock URL \url{https://cds.cern.ch/record/592033}.

\bibitem[Fong and Spivak(2018)]{fong2018seven}
Brendan Fong and David~I Spivak.
\newblock \emph{Seven Sketches in Compositionality: An Invitation to Applied
  Category Theory}.
\newblock 2018.
\newblock URL \url{http://arxiv.org/abs/1803.05316}.
\newblock cite arxiv:1803.05316Comment: 341+xii pages.

\bibitem[Spirtes et~al.(2000)Spirtes, Glymour, and Scheines]{spirtes:book}
Peter Spirtes, Clark Glymour, and Richard Scheines.
\newblock \emph{Causation, Prediction, and Search, Second Edition}.
\newblock Adaptive computation and machine learning. {MIT} Press, 2000.
\newblock ISBN 978-0-262-19440-2.

\bibitem[Lewis(1973)]{DBLP:journals/jphil/Lewis73}
David Lewis.
\newblock Counterfactuals and comparative possibility.
\newblock \emph{J. Philos. Log.}, 2\penalty0 (4):\penalty0 418--446, 1973.
\newblock \doi{10.1007/BF00262950}.
\newblock URL \url{https://doi.org/10.1007/BF00262950}.

\bibitem[Halpern(2016)]{halpern:ac}
Joseph~Y. Halpern.
\newblock \emph{Actual Causality}.
\newblock {MIT} Press, 2016.
\newblock ISBN 978-0-262-03502-6.

\bibitem[{de Araujo Fernandes} and Haeusler(2009)]{topos-counterfactual}
Ricardo~Queiroz {de Araujo Fernandes} and Edward~Hermann Haeusler.
\newblock A topos-theoretic approach to counterfactual logic.
\newblock \emph{Electronic Notes in Theoretical Computer Science},
  256:\penalty0 33--47, 2009.
\newblock ISSN 1571-0661.
\newblock \doi{https://doi.org/10.1016/j.entcs.2009.11.004}.
\newblock URL
  \url{https://www.sciencedirect.com/science/article/pii/S157106610900454X}.
\newblock Proceedings of the Fourth Workshop on Logical and Semantic
  Frameworks, with Applications (LSFA 2009).

\bibitem[Baez and Stay(2010)]{Baez_2010}
J.~Baez and M.~Stay.
\newblock Physics, topology, logic and computation: A rosetta stone.
\newblock In \emph{New Structures for Physics}, pages 95--172. Springer Berlin
  Heidelberg, 2010.
\newblock \doi{10.1007/978-3-642-12821-9_2}.
\newblock URL \url{https://doi.org/10.1007%2F978-3-642-12821-9_2}.

\bibitem[Fong(2012)]{fong:ms}
Brendan Fong.
\newblock Causal theories: A categorical perspective on bayesian networks,
  2012.

\bibitem[Pearl(2009{\natexlab{b}})]{pearl-book}
Judea Pearl.
\newblock \emph{Causality: Models, Reasoning and Inference}.
\newblock Cambridge University Press, USA, 2nd edition, 2009{\natexlab{b}}.
\newblock ISBN 052189560X.

\bibitem[Barmak(2011)]{barmak}
Jonathan~A. Barmak.
\newblock \emph{Algebraic topology of finite topological spaces and
  applications}.
\newblock Lecture notes in mathematics <Berlin>. Springer, Heidelberg ; Berlin
  u.a., 2011.
\newblock URL
  \url{http://deposit.d-nb.de/cgi-bin/dokserv?id=3826587&prov=M&dok%5Fvar=1&dok%5Fext=htm}.

\bibitem[Joyal et~al.(1996)Joyal, Nielsen, and Winskel]{JOYAL1996164}
Andre Joyal, Mogens Nielsen, and Glynn Winskel.
\newblock Bisimulation from open maps.
\newblock \emph{Information and Computation}, 127\penalty0 (2):\penalty0
  164--185, 1996.
\newblock ISSN 0890-5401.
\newblock \doi{https://doi.org/10.1006/inco.1996.0057}.
\newblock URL
  \url{https://www.sciencedirect.com/science/article/pii/S0890540196900577}.

\bibitem[Mahadevan(2021{\natexlab{b}})]{sm:udm}
Sridhar Mahadevan.
\newblock Universal decision models.
\newblock \emph{CoRR}, abs/2110.15431, 2021{\natexlab{b}}.
\newblock URL \url{https://arxiv.org/abs/2110.15431}.

\bibitem[Nagurney(1999)]{nagurney:vibook}
A.~Nagurney.
\newblock \emph{Network Economics: A Variational Inequality Approach}.
\newblock Kluwer Academic Press, 1999.

\bibitem[Hansen and Ghrist(2020)]{discourse-sheaves}
Jakob Hansen and Robert Ghrist.
\newblock Opinion dynamics on discourse sheaves, 2020.
\newblock URL \url{https://arxiv.org/abs/2005.12798}.

\bibitem[Forre and Mooij(2017)]{hedge}
Patrick Forre and Joris~M. Mooij.
\newblock Markov properties for graphical models with cycles and latent
  variables, 2017.

\bibitem[Aristotle(1984)]{aristotle}
Aristotle.
\newblock \emph{The Complete Works of Aristotle, Volumes 1 and 2}.
\newblock Princeton University Press, 1984.

\bibitem[Descartes(1644)]{descartes}
Rene Descartes.
\newblock \emph{Principles of Philosophy}.
\newblock Dordecht: Reidel, 1644.

\bibitem[Hume(1740)]{hume}
David Hume.
\newblock \emph{A Treatise of Human Nature}.
\newblock Oxford University Press, 1740.

\bibitem[Darwin(1876)]{darwin:1876}
Charles Darwin.
\newblock \emph{The effect of cross and self-fertilization in the vegetable
  kingdom}.
\newblock John Murray, 1876.

\bibitem[Fisher(1925)]{fisher}
R.A. Fisher.
\newblock \emph{Statistical Methods for Research Workers}.
\newblock 1925.

\bibitem[Wright(1921)]{Wright1921CorrelationAndCausation}
Sewall Wright.
\newblock Correlation and causation.
\newblock \emph{Journal of agricultural research}, 20\penalty0 (7):\penalty0
  557--585, 1921.

\bibitem[Evans(2018)]{mdag}
Robin~J. Evans.
\newblock {Margins of discrete Bayesian networks}.
\newblock \emph{The Annals of Statistics}, 46\penalty0 (6A):\penalty0 2623 --
  2656, 2018.
\newblock \doi{10.1214/17-AOS1631}.
\newblock URL \url{https://doi.org/10.1214/17-AOS1631}.

\bibitem[Lauritzen(1996)]{lauritzen:text}
S.~Lauritzen.
\newblock \emph{Graphical Models}.
\newblock Oxford University Press, 1996.

\bibitem[Lauritzen and Richardson(2002)]{lauritzen:chain}
Steffen~L. Lauritzen and Thomas~S. Richardson.
\newblock Chain graph models and their causal interpretations.
\newblock \emph{Journal of the Royal Statistical Society: Series B (Statistical
  Methodology)}, 64\penalty0 (3):\penalty0 321--348, 2002.
\newblock \doi{https://doi.org/10.1111/1467-9868.00340}.
\newblock URL
  \url{https://rss.onlinelibrary.wiley.com/doi/abs/10.1111/1467-9868.00340}.

\bibitem[Eberhardt(2008)]{DBLP:conf/uai/Eberhardt08}
Frederick Eberhardt.
\newblock Almost optimal intervention sets for causal discovery.
\newblock In David~A. McAllester and Petri Myllym{\"{a}}ki, editors,
  \emph{{UAI} 2008, Proceedings of the 24th Conference in Uncertainty in
  Artificial Intelligence, Helsinki, Finland, July 9-12, 2008}, pages 161--168.
  {AUAI} Press, 2008.
\newblock URL
  \url{https://dslpitt.org/uai/displayArticleDetails.jsp?mmnu=1\&smnu=2\&article\_id=1948\&proceeding\_id=24}.

\bibitem[Hauser and
  B{\"{u}}hlmann(2012{\natexlab{a}})]{DBLP:journals/jmlr/HauserB12}
Alain Hauser and Peter B{\"{u}}hlmann.
\newblock Characterization and greedy learning of interventional markov
  equivalence classes of directed acyclic graphs.
\newblock \emph{J. Mach. Learn. Res.}, 13:\penalty0 2409--2464,
  2012{\natexlab{a}}.
\newblock URL \url{http://dl.acm.org/citation.cfm?id=2503320}.

\bibitem[Kocaoglu et~al.(2017)Kocaoglu, Shanmugam, and
  Bareinboim]{DBLP:conf/nips/KocaogluSB17}
Murat Kocaoglu, Karthikeyan Shanmugam, and Elias Bareinboim.
\newblock Experimental design for learning causal graphs with latent variables.
\newblock In Isabelle Guyon, Ulrike von Luxburg, Samy Bengio, Hanna~M. Wallach,
  Rob Fergus, S.~V.~N. Vishwanathan, and Roman Garnett, editors, \emph{Advances
  in Neural Information Processing Systems 30: Annual Conference on Neural
  Information Processing Systems 2017, December 4-9, 2017, Long Beach, CA,
  {USA}}, pages 7018--7028, 2017.
\newblock URL
  \url{https://proceedings.neurips.cc/paper/2017/hash/291d43c696d8c3704cdbe0a72ade5f6c-Abstract.html}.

\bibitem[Mao-cheng(1984)]{MAOCHENG198415}
CAI Mao-cheng.
\newblock On separating systems of graphs.
\newblock \emph{Discrete Mathematics}, 49\penalty0 (1):\penalty0 15--20, 1984.
\newblock ISSN 0012-365X.
\newblock \doi{https://doi.org/10.1016/0012-365X(84)90146-8}.
\newblock URL
  \url{https://www.sciencedirect.com/science/article/pii/0012365X84901468}.

\bibitem[Tadepalli and Russell(2021)]{prasad:aaai21}
Prasad Tadepalli and Stuart Russell.
\newblock {PAC} learning of causal trees with latent variables.
\newblock In \emph{AAAI}, 2021.

\bibitem[Shpitser et~al.(2010)Shpitser, VanderWeele, and
  Robins]{DBLP:conf/uai/ShpitserVR10}
Ilya Shpitser, Tyler~J. VanderWeele, and James~M. Robins.
\newblock On the validity of covariate adjustment for estimating causal
  effects.
\newblock In Peter Gr{\"{u}}nwald and Peter Spirtes, editors, \emph{{UAI} 2010,
  Proceedings of the Twenty-Sixth Conference on Uncertainty in Artificial
  Intelligence, Catalina Island, CA, USA, July 8-11, 2010}, pages 527--536.
  {AUAI} Press, 2010.
\newblock URL
  \url{https://dslpitt.org/uai/displayArticleDetails.jsp?mmnu=1\&smnu=2\&article\_id=2078\&proceeding\_id=26}.

\bibitem[Imbens and Rubin(2015)]{rubin-book}
Guido~W. Imbens and Donald~B. Rubin.
\newblock \emph{Causal Inference for Statistics, Social, and Biomedical
  Sciences: An Introduction}.
\newblock Cambridge University Press, USA, 2015.
\newblock ISBN 0521885884.

\bibitem[Maiti(2011)]{DBLP:reference/stat/Maiti11}
Tapabrata Maiti.
\newblock Horvitz-thompson estimator.
\newblock In Miodrag Lovric, editor, \emph{International Encyclopedia of
  Statistical Science}, pages 637--638. Springer, 2011.
\newblock \doi{10.1007/978-3-642-04898-2\_291}.
\newblock URL \url{https://doi.org/10.1007/978-3-642-04898-2\_291}.

\bibitem[Harshaw et~al.(2019)Harshaw, S{\"{a}}vje, Spielman, and
  Zhang]{gs-random-walk}
Christopher Harshaw, Fredrik S{\"{a}}vje, Daniel Spielman, and Peng Zhang.
\newblock Balancing covariates in randomized experiments using the gram-schmidt
  walk.
\newblock \emph{CoRR}, abs/1911.03071, 2019.
\newblock URL \url{http://arxiv.org/abs/1911.03071}.

\bibitem[Bansal et~al.(2020)Bansal, Jiang, Singla, and
  Sinha]{DBLP:conf/stoc/BansalJ0S20}
Nikhil Bansal, Haotian Jiang, Sahil Singla, and Makrand Sinha.
\newblock Online vector balancing and geometric discrepancy.
\newblock In Konstantin Makarychev, Yury Makarychev, Madhur Tulsiani, Gautam
  Kamath, and Julia Chuzhoy, editors, \emph{Proccedings of the 52nd Annual
  {ACM} {SIGACT} Symposium on Theory of Computing, {STOC} 2020, Chicago, IL,
  USA, June 22-26, 2020}, pages 1139--1152. {ACM}, 2020.
\newblock \doi{10.1145/3357713.3384280}.
\newblock URL \url{https://doi.org/10.1145/3357713.3384280}.

\bibitem[McInnes et~al.(2018)McInnes, Healy, and Melville]{mcinnes2018uniform}
Leland McInnes, John Healy, and James Melville.
\newblock Umap: Uniform manifold approximation and projection for dimension
  reduction, 2018.
\newblock URL \url{http://arxiv.org/abs/1802.03426}.
\newblock cite arxiv:1802.03426Comment: Reference implementation available at
  http://github.com/lmcinnes/umap.

\bibitem[Carlsson and M{\'{e}}moli(2013)]{DBLP:journals/focm/CarlssonM13}
Gunnar~E. Carlsson and Facundo M{\'{e}}moli.
\newblock Classifying clustering schemes.
\newblock \emph{Found. Comput. Math.}, 13\penalty0 (2):\penalty0 221--252,
  2013.
\newblock \doi{10.1007/s10208-012-9141-9}.
\newblock URL \url{https://doi.org/10.1007/s10208-012-9141-9}.

\bibitem[Reynolds(2009)]{DBLP:conf/esop/Reynolds09}
John~C. Reynolds.
\newblock Using category theory to design programming languages.
\newblock In Giuseppe Castagna, editor, \emph{Programming Languages and
  Systems, 18th European Symposium on Programming, {ESOP} 2009, Held as Part of
  the Joint European Conferences on Theory and Practice of Software, {ETAPS}
  2009, York, UK, March 22-29, 2009. Proceedings}, volume 5502 of \emph{Lecture
  Notes in Computer Science}, pages 62--63. Springer, 2009.
\newblock \doi{10.1007/978-3-642-00590-9\_5}.
\newblock URL \url{https://doi.org/10.1007/978-3-642-00590-9\_5}.

\bibitem[Alexandroff()]{alexandroff:1937}
P.~S. Alexandroff.
\newblock \emph{{D}iskrete {R}\"aume}.
\newblock \emph{Rec. Math. [Mat. Sbornik] N.S.}, 2:\penalty0 501--518.

\bibitem[Baez and Fritz(2014)]{entropy-as-a-functor}
John~C. Baez and Tobias Fritz.
\newblock A bayesian characterization of relative entropy.
\newblock 2014.
\newblock \doi{10.48550/ARXIV.1402.3067}.
\newblock URL \url{https://arxiv.org/abs/1402.3067}.

\bibitem[Janzing et~al.(2013)Janzing, Balduzzi, Grosse-Wentrup, and
  Scholkopf]{janzing}
Dominik Janzing, David Balduzzi, Moritz Grosse-Wentrup, and Bernhard
  Scholkopf.
\newblock {Quantifying causal influences}.
\newblock \emph{The Annals of Statistics}, 41\penalty0 (5):\penalty0 2324 --
  2358, 2013.
\newblock \doi{10.1214/13-AOS1145}.
\newblock URL \url{https://doi.org/10.1214/13-AOS1145}.

\bibitem[Zigler and Papadogeorgou(2018)]{zigler2018bipartite}
Corwin~M. Zigler and Georgia Papadogeorgou.
\newblock Bipartite causal inference with interference, 2018.

\bibitem[Adam and Dahleh(2019)]{adams}
Elie~M. Adam and Munther~A. Dahleh.
\newblock Generativity and interactional effects: an overview, 2019.
\newblock URL \url{https://arxiv.org/abs/1911.10406}.

\bibitem[Mahadevan(2021{\natexlab{c}})]{sm:causal-network-econ}
Sridhar Mahadevan.
\newblock Causal inference in network economics.
\newblock \emph{CoRR}, abs/2109.11344, 2021{\natexlab{c}}.
\newblock URL \url{https://arxiv.org/abs/2109.11344}.

\bibitem[Bello and Honorio(2018)]{DBLP:conf/nips/BelloH18a}
Kevin Bello and Jean Honorio.
\newblock Computationally and statistically efficient learning of causal bayes
  nets using path queries.
\newblock In Samy Bengio, Hanna~M. Wallach, Hugo Larochelle, Kristen Grauman,
  Nicol{\`{o}} Cesa{-}Bianchi, and Roman Garnett, editors, \emph{Advances in
  Neural Information Processing Systems 31: Annual Conference on Neural
  Information Processing Systems 2018, NeurIPS 2018, December 3-8, 2018,
  Montr{\'{e}}al, Canada}, pages 10954--10964, 2018.
\newblock URL
  \url{https://proceedings.neurips.cc/paper/2018/hash/a0b45d1bb84fe1bedbb8449764c4d5d5-Abstract.html}.

\bibitem[Katona(1966)]{sepsets}
Gyula Katona.
\newblock {On separating systems of a finite set}.
\newblock \emph{Journal of Combinatorial Theory}, 2\penalty0 (1):\penalty0
  174--194, 1966.

\bibitem[Hauser and
  B{\"{u}}hlmann(2012{\natexlab{b}})]{DBLP:journals/corr/hauser-arxiv}
Alain Hauser and Peter B{\"{u}}hlmann.
\newblock Two optimal strategies for active learning of causal models from
  interventions.
\newblock \emph{CoRR}, abs/1205.4174, 2012{\natexlab{b}}.
\newblock URL \url{http://arxiv.org/abs/1205.4174}.

\bibitem[Greinacher et~al.(2021)Greinacher, Thiele, Warkentin, Weisser, Kyrle,
  and Eichinger]{doi:10.1056/NEJMoa2104840}
Andreas Greinacher, Thomas Thiele, Theodore~E. Warkentin, Karin Weisser,
  Paul~A. Kyrle, and Sabine Eichinger.
\newblock Thrombotic thrombocytopenia after chadox1 ncov-19 vaccination.
\newblock \emph{New England Journal of Medicine}, 2021.
\newblock \doi{10.1056/NEJMoa2104840}.
\newblock URL \url{https://doi.org/10.1056/NEJMoa2104840}.

\bibitem[Schultz et~al.(2021)Schultz, S√∏rvoll, Michelsen, Munthe,
  Lund-Johansen, Ahlen, Wiedmann, Aamodt, Skatt√∏r, Tj√∏nnfjord, and
  Holme]{doi:10.1056/NEJMoa2104882}
Nina~H. Schultz, Ingvild~H. Sorvoll, Annika~E. Michelsen, Ludvig~A. Munthe,
  Fridtjof Lund-Johansen, Maria~T. Ahlen, Markus Wiedmann, Anne-Hege Aamodt,
  Thor~H. Skattor, Geir~E. Tjonnfjord, and Pal A. Holme.
\newblock Thrombosis and thrombocytopenia after chadox1 ncov-19 vaccination.
\newblock \emph{New England Journal of Medicine}, 2021.
\newblock \doi{10.1056/NEJMoa2104882}.
\newblock URL \url{https://doi.org/10.1056/NEJMoa2104882}.

\bibitem[Selinger(2010)]{Selinger_2010}
P.~Selinger.
\newblock A survey of graphical languages for monoidal categories.
\newblock In \emph{New Structures for Physics}, pages 289--355. Springer Berlin
  Heidelberg, 2010.
\newblock \doi{10.1007/978-3-642-12821-9_4}.
\newblock URL \url{https://doi.org/10.1007%2F978-3-642-12821-9_4}.

\bibitem[Baez and Erbele(2015)]{baez2015categories}
John~C. Baez and Jason Erbele.
\newblock Categories in control, 2015.

\bibitem[Censi(2015)]{censi}
Andrea Censi.
\newblock A mathematical theory of co-design, 2015.
\newblock URL \url{https://arxiv.org/abs/1512.08055}.

\bibitem[Fong(2015)]{cospans}
Brendan Fong.
\newblock Decorated cospans.
\newblock 2015.
\newblock \doi{10.48550/ARXIV.1502.00872}.
\newblock URL \url{https://arxiv.org/abs/1502.00872}.

\bibitem[Leinster(2003)]{higher-operads}
Tom Leinster.
\newblock Higher operads, higher categories, 2003.
\newblock URL \url{https://arxiv.org/abs/math/0305049}.

\bibitem[Spivak(2013)]{operads-db}
David~I. Spivak.
\newblock The operad of wiring diagrams: formalizing a graphical language for
  databases, recursion, and plug-and-play circuits, 2013.
\newblock URL \url{https://arxiv.org/abs/1305.0297}.

\bibitem[Joyal et~al.(1993)Joyal, Nielsen, and
  Winskel]{DBLP:conf/lics/JoyalNW93}
Andr{\'{e}} Joyal, Mogens Nielsen, and Glynn Winskel.
\newblock Bisimulation and open maps.
\newblock In \emph{Proceedings of the Eighth Annual Symposium on Logic in
  Computer Science {(LICS} '93), Montreal, Canada, June 19-23, 1993}, pages
  418--427. {IEEE} Computer Society, 1993.
\newblock \doi{10.1109/LICS.1993.287566}.
\newblock URL \url{https://doi.org/10.1109/LICS.1993.287566}.

\bibitem[Cattani and Winskel(2005)]{bisim}
Gian~Luca Cattani and Glynn Winskel.
\newblock Profunctors, open maps and bisimulation.
\newblock \emph{Math. Struct. Comput. Sci.}, 15\penalty0 (3):\penalty0
  553--614, 2005.
\newblock \doi{10.1017/S0960129505004718}.
\newblock URL \url{https://doi.org/10.1017/S0960129505004718}.

\bibitem[Dyson(2009)]{dyson:birds-and-frogs}
Freeman Dyson.
\newblock Birds and frogs.
\newblock \emph{American Mathematical Society}, 56\penalty0 (2):\penalty0
  212--223, 2009.
\newblock URL \url{https://www.ams.org/notices/200902/rtx090200212p.pdf}.

\end{thebibliography}

\end{document}